\documentclass{article}

\usepackage{arxiv}

\usepackage[utf8]{inputenc} 
\usepackage[T1]{fontenc}    
\usepackage{hyperref}       
\usepackage{nicefrac}       
\usepackage{microtype}      
\usepackage{lipsum}         
\usepackage{natbib}
\usepackage{doi}

\usepackage{amssymb}
\usepackage{makecell}
\usepackage{multirow}
\usepackage{booktabs}
\usepackage{subcaption}
\usepackage{amsmath,amsfonts}
\usepackage{algorithmic}
\usepackage{array}
\usepackage{textcomp}
\usepackage{stfloats}
\usepackage{url}
\usepackage{verbatim}
\usepackage{graphicx}
\usepackage{balance}

\usepackage{amsthm,mathtools}

\newcommand{\MI}{\mathcal{I}}   
\newcommand{\Id}{\mathbf{I}}  

\title{Exchange Is All You Need for Remote Sensing Change Detection}

\newif\ifuniqueAffiliation
\uniqueAffiliationtrue
\ifuniqueAffiliation 
\author{
	\href{https://orcid.org/0000-0003-4218-6790}{\includegraphics[scale=0.06]{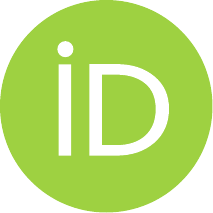}\hspace{1mm}Sijun~Dong} \\
	Wuhan University\\
	\And
	{\includegraphics[scale=0.06]{orcid.pdf}\hspace{1mm}Siming~Fu} \\
	Wuhan University\\
	\And
	{\includegraphics[scale=0.06]{orcid.pdf}\hspace{1mm}Kaiyu~Li} \\
	Xi'an Jiaotong University\\
	\And
	{\includegraphics[scale=0.06]{orcid.pdf}\hspace{1mm}Xiangyong~Cao} \\
	Xi'an Jiaotong University\\
	\And
	{\includegraphics[scale=0.06]{orcid.pdf}\hspace{1mm}Xiaoliang~Meng}\thanks{Corresponding author: \texttt{xmeng@whu.edu.cn}} \\
	Wuhan University\\
	\And
	{\includegraphics[scale=0.06]{orcid.pdf}\hspace{1mm}Bo~Du} \\
	Wuhan University
}
\else
\fi

\renewcommand{\shorttitle}{\textit{arXiv} Template}

\hypersetup{
pdftitle={Exchange Is All You Need for Remote Sensing Change Detection},
pdfsubject={q-bio.NC, q-bio.QM},
pdfauthor={Sijun Dong, Siming Fu, Kaiyu Li, Xiangyong Cao, Xiaoliang Meng, Bo Du},
pdfkeywords={Remote sensing change detection, Feature exchange, Siamese Encoder-Exchange-Decoder, Mutual information invariance, Pixel consistency},
}

\begin{document}
\maketitle

\begin{abstract}
	Remote sensing change detection fundamentally relies on the effective fusion and discrimination of bi-temporal features. Prevailing paradigms typically utilize Siamese encoders bridged by explicit difference-computation modules (e.g., subtraction or concatenation) to identify changes. In this work, we challenge this complexity with SEED (Siamese Encoder–Exchange–Decoder), a streamlined paradigm that replaces explicit differencing with parameter-free feature exchange. By sharing weights across both Siamese encoders and decoders, SEED effectively operates as a single-parameter-set model. Theoretically, we formalize feature exchange as an orthogonal permutation operator; we prove that, under pixel consistency, this mechanism preserves Mutual Information and Bayes-optimal risk, whereas common arithmetic fusion methods often introduce information loss. Extensive experiments across five benchmarks (SYSU-CD, LEVIR-CD, PX-CLCD, WaterCD, CDD) and three backbones (SwinT, EfficientNet, ResNet) demonstrate that SEED matches or surpasses state-of-the-art methods despite its simplicity. Furthermore, we reveal that standard semantic segmentation models can be transformed into competitive change detectors solely by inserting this exchange mechanism (SEG2CD). The proposed paradigm offers a robust, unified, and interpretable framework for change detection, proving that simple feature exchange is sufficient for high-performance information fusion. Code and full training/evaluation protocols will be released at \url{https://github.com/dyzy41/open-rscd}.
\end{abstract}

\keywords{Remote sensing change detection \and Feature exchange \and Siamese Encoder-Exchange-Decoder \and Mutual information invariance \and Pixel consistency}

\section{Introduction}

Remote sensing change detection aims to identify differences between two images of the same geographic area acquired at different times. It underpins applications such as land-use monitoring, urban expansion analysis, ecological protection, and disaster assessment~\cite{ting_bai_deep_2023}. In computer vision, change detection and semantic segmentation are both pixel-level prediction tasks that typically adopt an encoder-decoder architecture: the encoder maps inputs to high-dimensional features, and the decoder reconstructs them into the task-specific output. Unlike single-temporal segmentation, change detection must process \emph{bi-temporal} inputs and therefore commonly relies on a Siamese encoder with shared weights to maintain representation alignment~\cite{dong_efficientcd_2024}. As a result, most contemporary change-detection methods are deep learning--based~\cite{chen_remote_2022, zhang_swinsunet_2022, wang2022unetformer}.

\begin{figure*}[!t]
  \centering
  \includegraphics[width=\textwidth]{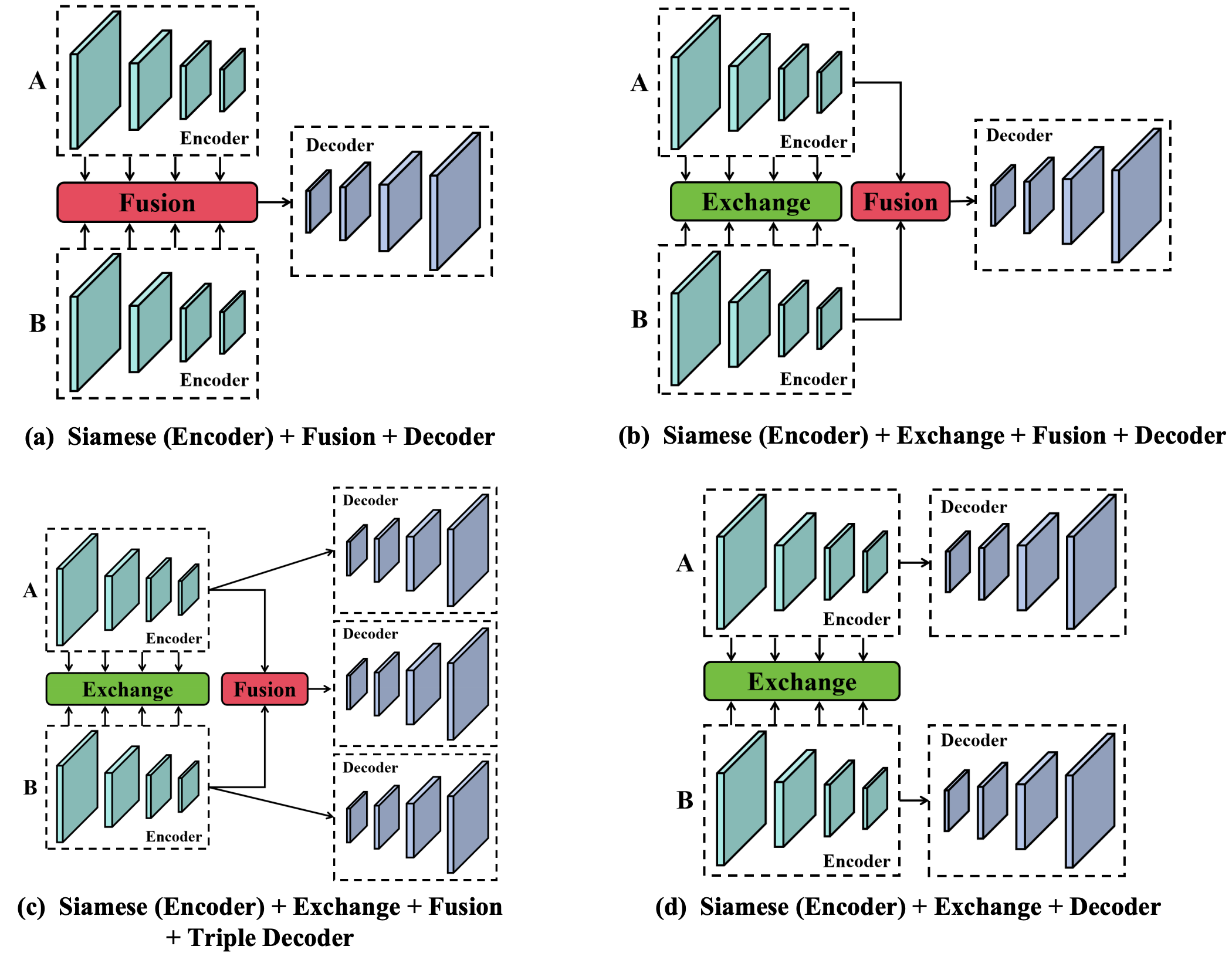}
  \caption{Comparison of change-detection paradigms. (a) A Siamese encoder extracts bi-temporal features, which are then fused and decoded by a single branch. (b) Feature exchange is performed during encoding to enhance cross-temporal interaction, followed by fusion and single-branch decoding. (c) As in (b), but in addition to the fused branch, each exchanged branch is decoded separately (a triple-decoder). (d) Exchange without explicit fusion: each exchanged encoder branch is decoded separately.}
  \label{fig:change_detection_frameworks}
\end{figure*}

In mainstream deep learning-based change detection, models first extract features from the two times and then construct \emph{difference features}—via addition/subtraction, concatenation, or dedicated modules—to drive a single decoder that localizes changes~\cite{shi_change_2020,lv_land_2022}. Siamese networks provide an effective mechanism for feature extraction and alignment in this pipeline~\cite{shi_deeply_2022}: weight-sharing encoders obtain bi-temporal representations from which difference features are built and subsequently decoded~\cite{zhu_land-useland-cover_2022-1}. Building on this line, we next discuss a fusion-free alternative that avoids explicit differencing.

\begin{figure*}[!t]
  \centering
  \includegraphics[width=\textwidth]{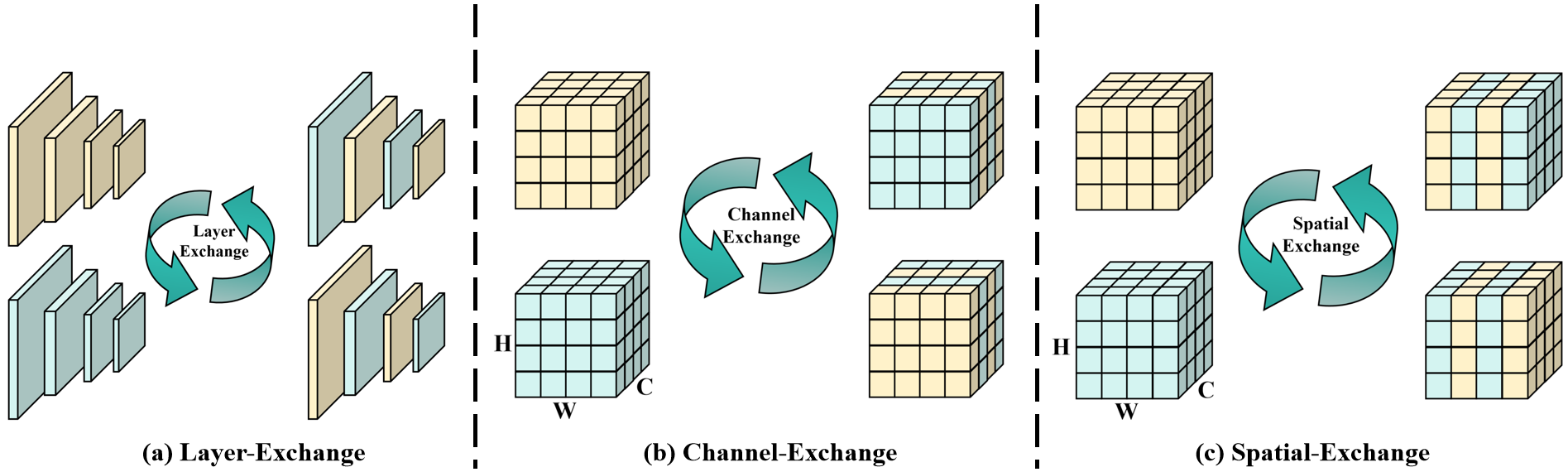}
  \caption{Feature exchange methods in remote sensing change detection.}
  \label{fig:seed_exchange}
\end{figure*}

Another widely studied aspect of change detection is the computation of difference features~\cite{pei_feature_2022,x_zhang_difunet_2022, gu2023fdff}. Since change detection models take two images as input but produce only a single output, it is necessary to fuse the features of these two images within the model. In common change detection models, there are three simple methods to fuse bi-temporal features (fusion Layer): concatenation~\cite{gu2023fdff,wan_d-tnet_2022-3, wang_hmcnet_2022-2}, element-wise addition~\cite{pei_feature_2022, gu2023fdff}, and element-wise subtraction~\cite{gu2023fdff, shi_deeply_2022, x_zhang_difunet_2022}. The concatenation method stacks the bi-temporal feature maps along the channel dimension so that the model can learn the change features from the stacked maps; the addition method, a common feature fusion technique in deep learning, performs element-wise summation of the two feature maps to enable the model to extract change information from the combined features; while the subtraction method directly computes the element-wise difference to represent the change features.

However, arithmetic fusion can fundamentally distort the original representations: except for pure concatenation without subsequent reduction, addition and subtraction are non-invertible linear maps that may discard information and worsen numerical conditioning, thereby complicating optimization. 

To this end, researchers have proposed difference‐learning methods based on feature exchange, and have introduced various paradigms built on feature exchange, as shown in Figure~\ref{fig:change_detection_frameworks}. Among them, Figure~\ref{fig:change_detection_frameworks}(a) depicts the classic Siamese neural‐network–based change‐detection paradigm, on which most mainstream change‐detection algorithms are currently designed. The paradigm in Figure~\ref{fig:change_detection_frameworks}(b) is a feature‐exchange–based change‐detection framework that, during the encoding stage, uses feature exchange to strengthen information flow between the two temporal images. The paradigm in Figure~\ref{fig:change_detection_frameworks}(c) is similar to Figure~\ref{fig:change_detection_frameworks}(b), except that in the decoding stage the exchanged bi‐temporal feature maps are decoded separately, thereby constructing a triple‐branch decoder. From Figure~\ref{fig:change_detection_frameworks}(a), Figure~\ref{fig:change_detection_frameworks}(b), and Figure~\ref{fig:change_detection_frameworks}(c), it is evident that current change‐detection models still rely on a Fusion Layer to merge bi‐temporal features—because in traditional frameworks it is generally believed that bi‐temporal features must be fused to represent change features. However, in this paper we propose a novel Siamese Encoder‐Exchange‐Decoder (SEED) framework. As shown in Figure~\ref{fig:change_detection_frameworks}(d), SEED replaces explicit, parameter-heavy fusion with a parameter-free, information-preserving exchange mechanism, followed by decoding of the exchanged branches with shared weights. This yields a particularly concise pipeline compared with prior paradigms.

As for feature exchange methods, prior studies have proposed difference methods based on feature exchange to improve change detection models, which primarily include layer exchange~\cite{dong_efficientcd_2024}, channel exchange~\cite{fang_changer_2022, zhao_exchanging_2023}, and spatial exchange~\cite{fang_changer_2022, zhao_exchanging_2023}, as illustrated in Figure~\ref{fig:seed_exchange}. These methods utilize the three aforementioned feature exchange techniques to enhance the information exchange between bi-temporal images and thereby optimize the performance of change detection models. Nevertheless, they still rely on constructing difference features to learn the change regions, and a thorough analysis of the core principles of feature exchange remains lacking.

Based on previous studies~\cite{dong_efficientcd_2024, fang_changer_2022, zhao_exchanging_2023}, we identified a fundamental factor in change detection: as long as the correspondence of pixels between the bi-temporal images used for determining changes is preserved, the core objective of the change detection model remains unchanged—namely, to determine whether each corresponding pixel has changed in the target category. We refer to this principle as the pixel-consistency principle. Based on this insight, we propose a simple change detection framework, the Siamese Encoder-Exchange-Decoder (SEED) framework. SEED dispenses with explicit differencing or fusion and preserves the original features end-to-end. It uses \emph{parameter-free} exchange to mix information across times, and learns change characteristics from the exchanged Siamese branches with shared parameters.

In the proposed SEED framework, the Siamese encoders share the same encoding weights, the feature-exchange module introduces no additional parameters, and the Siamese decoders also share the same decoding weights. Consequently, the overall change-detection model effectively uses a \emph{single} encoder--decoder parameter set and dispenses with explicit fusion/differencing, offering a unified perspective that links change detection with semantic segmentation. This also makes it straightforward to convert most encoder--decoder segmentation models into SEED-based change-detection models. The main contributions of this paper can be summarized as follows:

\begin{itemize}

  \item We introduce an \emph{exchange-based} change-detection paradigm (SEED) that identifies change regions using a Siamese encoder--decoder with shared weights, \emph{without} computing explicit difference features.

  \item We provide a principled analysis of feature exchange by formalizing it as a permutation operator and, under \emph{pixel consistency}, proving mutual-information and Bayes-optimal risk invariance; this clarifies why common fusion/differencing (addition, subtraction, concatenation with reduction) can be information-losing.

  \item We conduct extensive experiments across five datasets and three backbones, with ablations on layer/channel/spatial (and random) exchange, demonstrating that SEED is simple, robust, and competitive or superior to strong baselines.

  \item For lightweight deployment, we show that single-decoder inference attains comparable accuracy while substantially reducing FLOPs. As supporting evidence of generality, we also demonstrate a simple SEG2CD recipe that converts standard segmentation models into competitive change-detection models by inserting exchange alone.

\end{itemize}

\section{Related Works}
\subsection{Change Detection Paradigms}
Due to the unique requirement of processing two input images in change detection tasks, the encoder-decoder architectures designed for change detection often take on different forms~\cite{dong_changeclip_2024}. In early research on change detection, the most common architecture was based on twin neural networks, where dual-branch encoders extract features from the bi-temporal images separately, and a difference feature pyramid is then constructed from these two feature maps~\cite{dong_efficientcd_2024, bandara_transformer-based_2022}. Finally, a single-branch decoder reconstructs the change map from the difference feature pyramid.

Subsequently, researchers proposed novel architectures specifically tailored for change detection. For example, Zhao et al. introduced the Triple-Stream Network~\cite{zhao_triple-stream_2023}. In addition to the two independent encoding branches for the bi-temporal images, the images are concatenated along the channel dimension to form a third encoding branch. Fang et al. proposed Changer~\cite{fang_changer_2022}, which promotes information exchange between the two encoding branches by performing feature exchange during the dual-branch encoding stage. Subsequently, the framework fuses the dual-branch feature pyramid and employs a single decoder to generate the final change map. Later, Zhao et al. introduced the SGSLN model~\cite{zhao_exchanging_2023}, which shares the same dual-branch and feature exchange encoding stage as Changer, but expands the decoder into three branches. One branch for fusing the bi-temporal feature maps and two separate branches for decoding each exchanged feature map. Therefore, SGSLN adopts a Siamese encoder and triple decoder architecture for change detection. Zhao et al. further argued that each exchanged feature map alone is sufficient to decode the change information which is a key insight that underpins our further exploration and optimization in the proposed SEED method.

\begin{figure*}[!t]
  \centering
  \includegraphics[width=\textwidth]{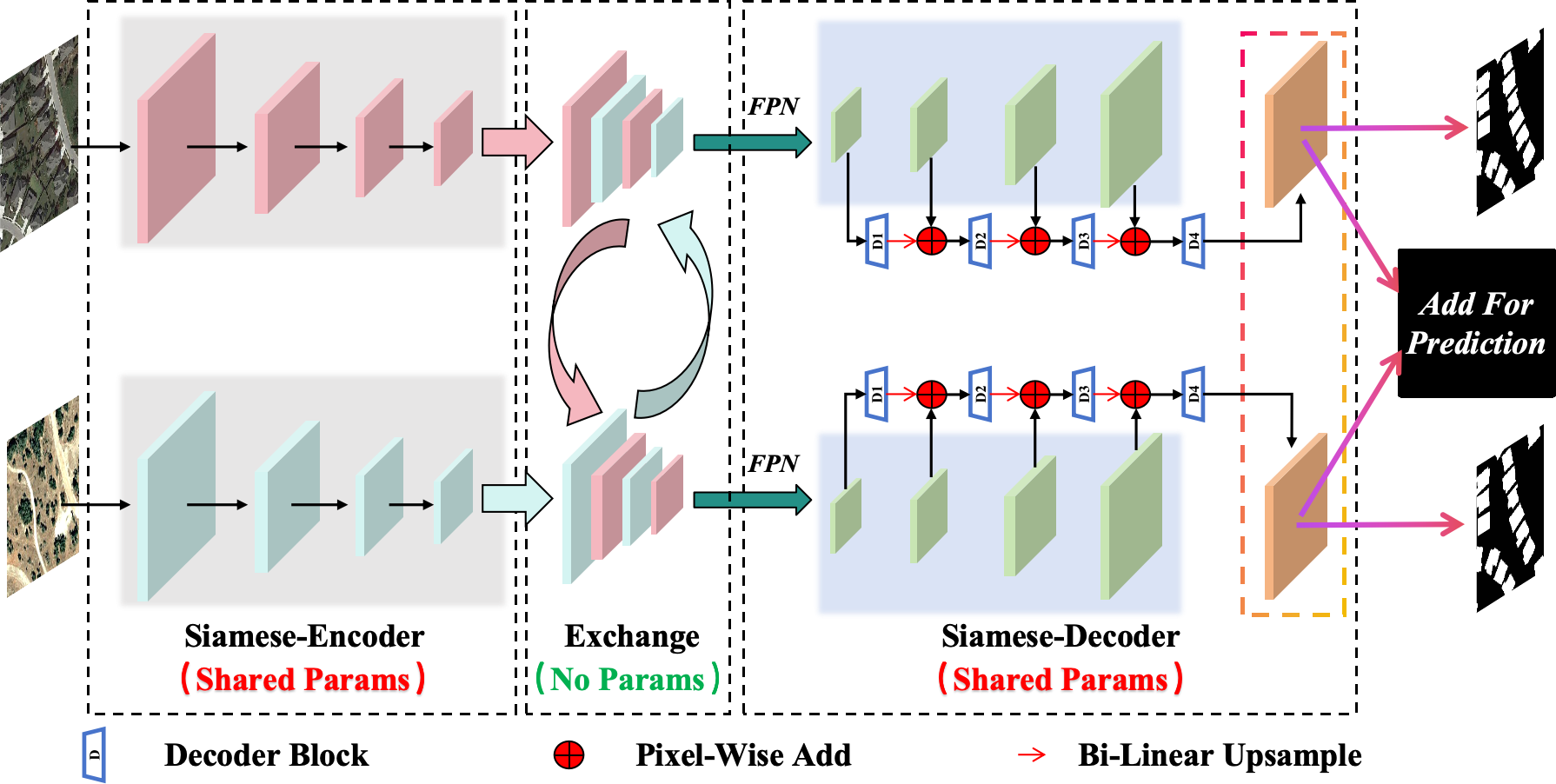}
  \caption{Siamese Encoder-Exchange-Decoder (SEED) Overall Architecture.}
  \label{fig:seed_framework}
\end{figure*}

Recently, the emergence of the Mamba architecture~\cite{dao_transformers_2024} has inspired several researchers to integrate the Mamba module into change detection models~\cite{zhang_cdmamba_2025, zhang_dc-mamba_2024, liu_cwmamba_2025}. By incorporating Mamba during the encoding phase, these methods enhance the model's ability to capture long-range dependencies, thereby improving the accuracy of change region detection. For instance, Chen et al. proposed ChangeMamba~\cite{chen_changemamba_2024}, a change detection framework based on Mamba, which adaptively captures global spatial dependencies within its dual-branch encoder and employs Mamba modules in the decoder to progressively fuse difference features, further enhancing the model's capability to learn discriminative change features. Similarly, RS-Mamba~\cite{zhao_rs-mamba_2024} applies the Mamba module to change detection tasks. In addition, extensive experiments conducted on remote sensing image segmentation by RS-Mamba have demonstrated that the Mamba architecture is highly effective for dense prediction tasks in remote sensing imagery.

\subsection{Difference Features In Change Detection}  
In most mainstream change detection models, the methods used to learn change regions typically rely on constructing difference features to represent changes~\cite{dong_efficientcd_2024, shi_deeply_2022, zhang_bifa_2024, noman2024remote, feng2023change, li_stade-cdnet_2024}. The existing approaches for computing difference features can be categorized into two classes: mathematical difference computation~\cite{dong_efficientcd_2024, shi_deeply_2022, feng2023change} and model-based learned difference computation~\cite{zhang_bifa_2024, noman2024remote, li_stade-cdnet_2024}.

In the mathematical methods, the simplest approach is to subtract one temporal image from the other to highlight the regions of change. Building on this idea, some researchers have devised more effective computational strategies. For example, in EfficientCD~\cite{dong_efficientcd_2024}, Dong et al. combine the Euclidean distance for pixel-level difference computation and use the resultant difference map as weights to recalibrate the original feature maps, thereby maximizing the prominence of change features. Feng et al. proposed DMINet~\cite{feng2023change}, a method that captures multi-scale differences by performing pixel subtraction in conjunction with channel-wise concatenation. They further utilize a simple Hadamard product to achieve lightweight feature aggregation from high to low levels, which facilitates precise localization and detailed capture of changes across different object scales.

\begin{figure*}[!t]
  \centering
  \includegraphics[width=\textwidth]{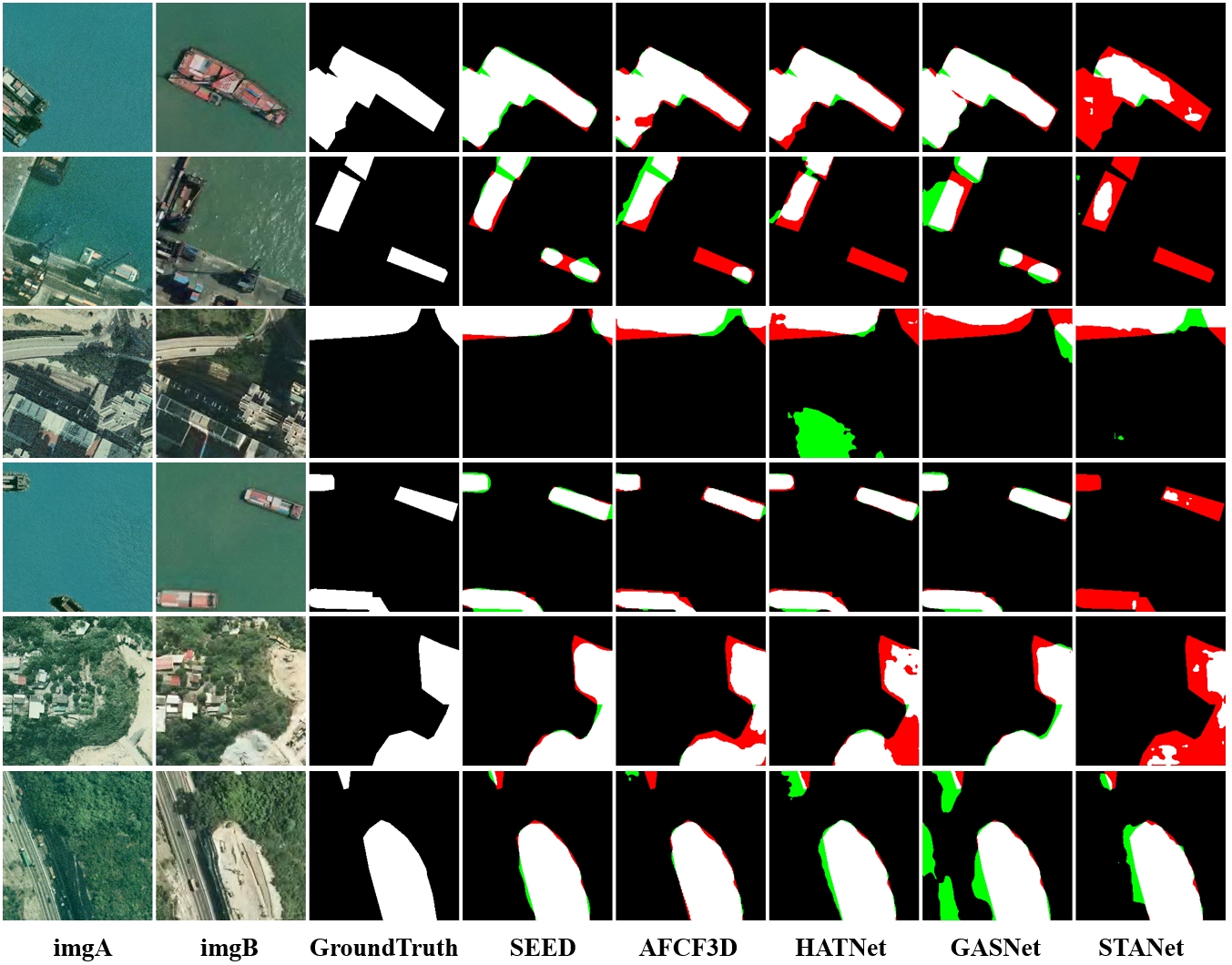}
  \caption{Visualization results in SYSU-CD dataset.}
  \label{fig:seed_sysu}
\end{figure*}

In model-based learned difference computation methods, researchers have designed dedicated modules to precisely represent change features. For example, in the ScratchFormer model, Noman et al.~\cite{noman2024remote} proposed a Change Enhancement Feature Fusion (CEFF) module. This module enhances the semantic discrepancies between the two feature pathways by re-weighting each channel at every scale and then outputs the enhanced features to the decoder to predict the final binary change map. Likewise, Li et al. proposed STADE-Net~\cite{li_stade-cdnet_2024} and introduced a Change Difference Enhancement Module (CDDM), which integrates additional feature extraction and data processing mechanisms to improve the accuracy of change detection. The CDDM is capable of effectively capturing the change information in the images while filtering out noise and redundant data, resulting in more reliable outcomes.

However, regardless of the method used for computing difference features, these approaches all involve processing the bi-temporal feature maps prior to feeding them into the decoder for change prediction. Such processing inevitably compromises the original semantic integrity of the bi-temporal features, causing the accuracy of the change detection to heavily depend on the efficacy of the difference feature extraction process.

\begin{table}[htbp]
\centering
\caption{Quantitative Results On The SYSU-CD Dataset}
\label{tab:sysu_cd_results}
\begin{tabular}{lccccc}
\hline
\textbf{Model} & \textbf{OA} & \textbf{IoU} & \textbf{F1} & \textbf{Rec} & \textbf{Prec} \\
\hline
STANet \cite{chen_spatial-temporal_2020} & 88.24 & 57.22 & 72.79 & 66.71 & 80.08 \\
DSAMNet \cite{shi_deeply_2022}    & --    & 64.18 & 78.18 & 81.86 & 74.81 \\
P2V \cite{lin_transition_2023}    & 90.49 & 66.29 & 79.73 & 79.29 & 80.17 \\
RCDT \cite{lu_cross_2024}         & --    & 67.46 & 80.57 & 86.21 & 75.62 \\
DARNet \cite{li_densely_2022}     & 91.26 & 68.10 & 81.03 & 79.11 & 83.04 \\
STDF-CD \cite{y_zhou_stdf_2025}   & 91.42 & 68.83 & 81.53 & 80.35 & 82.76 \\
STFF-GA \cite{h_wei_spatio-temporal_2024}   & --  & 69.45   & 81.97  & 80.14  & 83.89 \\
SGANet \cite{j_chen_sganet_2025}  & --    & 69.55 & 82.04 & 76.50 & 88.45 \\
MFCF-Net \cite{b_huang_remote-sensing_2024}   & 92.05   & 69.79   & 82.21   & 79.37   & 85.25 \\
AMFNet \cite{zhan_amfnet_2024}     & 92.30  & 69.85 &  82.25 &  82.51 &  88.23 \\
CMCD \cite{li_cmcd_2025}          & --    & 69.87 & 82.26 & 85.90 & 78.92 \\
MIFNet \cite{w_xie_mifnet_2025}       & --  & 70.09 & 82.42 & 79.47 & 85.85 \\
\hline
SEED (LE)                             & 92.03 & 70.33 & 82.58 & 80.16 & 85.16 \\
SEED (CE) & 92.16 & 70.91	& 82.98	& 81.01	& 85.05 \\
SEED (SE) & 91.52 & 69.05	& 81.69	& 80.27	& 83.16 \\
\hline
\end{tabular}
\end{table}

From the above analysis, it can be observed that optimization for change detection tasks mainly follows two complementary directions. First, one can design various model architectures that adapt to the bi-temporal input characteristics inherent to change detection. Secondly, one can develop effective learning modules specifically aimed at extracting and representing the difference features between the two temporal images. In this paper, we completely abandon any explicit modules for constructing difference features and instead adopt a pure encoder-decoder structure (SEED) to build the change detection model. The SEED framework frees change detection algorithms from the reliance on difference feature computation, thereby providing new and flexible design strategies.

\begin{figure*}[!t]
  \centering
  \includegraphics[width=\textwidth]{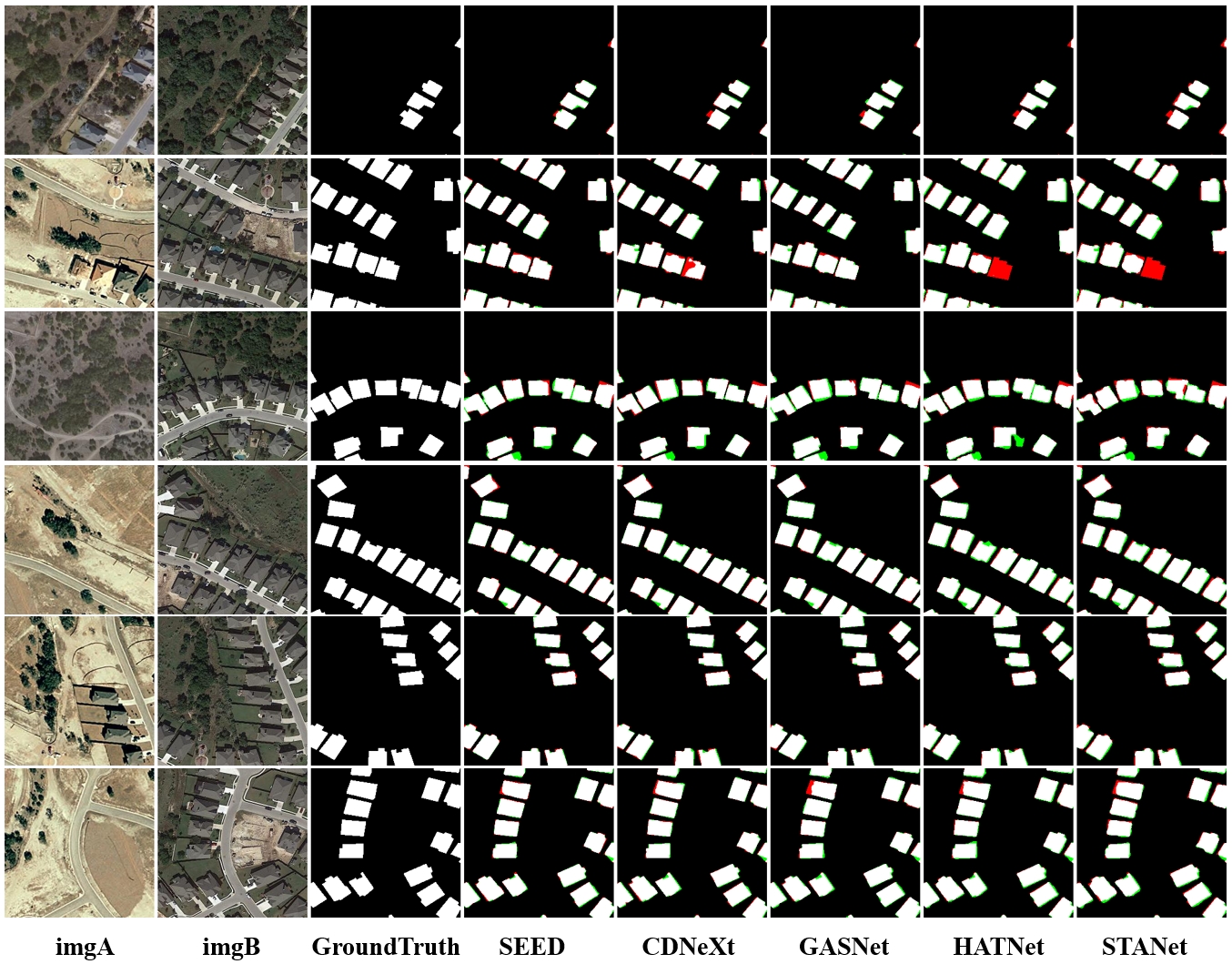}
  \caption{Visualization results in LEVIR-CD dataset.}
  \label{fig:seed_levir}
\end{figure*}

\begin{table}[htbp]
\centering
\caption{Quantitative Results On The LEVIR-CD Dataset}
\label{tab:levir_cd_results}
\begin{tabular}{lccccc}
\hline
\textbf{Model} & \textbf{OA} & \textbf{IoU} & \textbf{F1} & \textbf{Rec} & \textbf{Prec} \\
\hline
STANet \cite{chen_spatial-temporal_2020}   & 99.02 & 81.85 & 90.02 & 87.13 & 93.10 \\
CDMamba \cite{zhang_cdmamba_2025}          & 99.06 & 83.07 & 90.75 & 90.08 & 91.43 \\
AMFNet \cite{zhan_amfnet_2024}             & 99.07 & 83.13 & 90.79 & 91.15 & 94.77 \\
ISDANet \cite{h_ren_interactive_2025}      & 99.10 & 83.63 & 91.06 & --    & 92.19 \\
RSM-CD \cite{zhao_rs-mamba_2024}           & --    & 83.66 & 91.10 & 89.73 & 92.52 \\
DgFA \cite{f_zhou_dual-granularity_2025}   & 99.11 & 83.93 & 91.26 & 90.84 & 91.69 \\
TRSANet \cite{j_li_trsanet_2024}           & 99.14 & 84.22 & 91.44 & 89.95 & 92.97 \\
HATNet \cite{xu_hybrid_nodate}             & --    & 84.41 & 91.55 & 90.23 & 92.90 \\
GASNet \cite{zhang_global-aware_2023}      & 99.11 & --    & 91.21 & 90.62 & 91.82 \\
CMCD \cite{li_cmcd_2025}                   & --    & 84.50 & 91.60 & 90.48 & 92.74 \\
STFF-GA \cite{h_wei_spatio-temporal_2024}   & --  & 84.81   & 91.78  & 91.10  & 92.46 \\
MFCF-Net \cite{b_huang_remote-sensing_2024} & 99.17  & 84.82  & 91.79  & 90.84  & 92.76 \\
STDF-CD \cite{y_zhou_stdf_2025}            & 99.12 & 85.05 & 91.92 & 91.01 & 92.85 \\
SFFCE-CD \cite{y_xing_sffce-cd_2025}       & 99.20 & 85.32 & 92.08 & 91.18 & 93.00 \\
MIFNet \cite{w_xie_mifnet_2025}            & --  & 85.35 & 92.10 & 90.87 & 93.37 \\
SGANet \cite{j_chen_sganet_2025}           & --    & 85.45 & 92.14 & 91.03 & 93.30 \\
RCDT \cite{lu_cross_2024}                  & --    & 85.50 & 92.18 & 93.27 & 91.12 \\
FTA-Net \cite{t_zhu_fta-net_2025}          & --    & 85.58 & 92.23 & 92.68 & 91.79 \\
CDNeXt \cite{wei_robust_2024}              & 99.24 & 85.86 & 92.39 & 90.92 & 93.91 \\
HASNet \cite{c_tao_hasnet_2025}            & 99.53 & 85.90 & 92.42 & 92.04 & 92.80 \\
UA-BCD \cite{li_overcoming_2025}           & --    & 85.99 & 92.47 & 91.57 & 93.38 \\
FTransDF-Net \cite{li_dual_2025}           & --    & 86.04 & 92.50 & 91.40 & 93.62 \\
RSBuilding \cite{wang_rsbuilding_2024}     & --    & 86.19 & 92.59 & 91.80 & 93.39 \\
\hline
SEED (LE)                                      & 99.26 & 86.25 & 92.62 & 90.97 & 94.32 \\
SEED (CE) & 99.25	& 86.03	& 92.49	& 91.08	& 93.94 \\
SEED (SE) & 99.26	& 86.24	& 92.61	& 91.42	& 93.83 \\
\hline
\end{tabular}
\end{table}

\section{Method}
\subsection{Overall Architecture}  
This work primarily explores the framework design for change detection tasks~\cite{dong_efficientcd_2024, fang_changer_2022, zhao_exchanging_2023} and proposes the Siamese Encoder-Exchange-Decoder (SEED) framework, as illustrated in Figure~\ref{fig:seed_framework}. In the encoding stage of the SEED framework, we construct a Siamese encoder with shared weights using Swin Transformer V2-Base (SwinTv2)~\cite{liu_swin_2021-5}, EfficientNet-B4~\cite{tan_efficientnet_2019}, and ResNet50~\cite{he_deep_2016-1}. During the feature exchange stage, three distinct exchange methods are employed: feature map layer exchange, feature map channel exchange, and feature map spatial exchange. In the neck stage, a parameter-sharing Feature Pyramid Network (FPN)~\cite{lin_feature_2017} is utilized to process the bi-temporal feature pyramid that has undergone feature exchange. In the decoding stage, to fully validate the potential of the SEED framework, we construct an extremely simple layer-by-layer decoding structure with shared decoder parameters. Furthermore, for the SwinTv2 backbone, a simple Swin Transformer V2 Block is integrated within the decoder for feature optimization, whereas for EfficientNet and ResNet, the BottleNeck module from ResNet is employed to optimize the features. In the training stage, the two branches of the decoder calculate the loss separately. In the inference stage, we sum the two logits of the decoder and divide them by 2 as the inference result of the change detection model.

\subsection{Feature Exchange Methods}  
Building upon previous studies addressing the role of feature exchange in change detection tasks, this paper further investigates three feature exchange schemes in depth. Moreover, the experimental results of these three schemes effectively validate the pixel-consistency principle in change detection tasks. First, these feature exchange schemes promote the exchange of feature information between the bi-temporal images without introducing any additional learnable parameters, thereby enhancing the model’s understanding of the bi-temporal images. Second, the change detection framework based on these feature interaction schemes abandons the complex methods for computing difference features, enabling the model to learn the difference features between the bi-temporal images while preserving their original feature integrity. All three can be written as \emph{permutation operators} applied to paired features and therefore preserve information under the pixel-consistency principle. Below, we provide a detailed description of these three feature exchange schemes.

\subsubsection{Feature Map Layer-Exchange (LE)}
Feature map layer exchange performs feature exchange at the layer dimension, as shown in Figure~\ref{fig:seed_exchange} (a). Given two feature lists $\{x_i\}$ and $\{y_i\}$ from bi-temporal images, the corresponding features are exchanged based on a specified step length (e.g., every other layer). Specifically, assuming the feature lists are $\{x_0, x_1, x_2, x_3, \ldots\}$ and $\{y_0, y_1, y_2, y_3, \ldots\}$, a predefined step length (e.g., $2$) is used to exchange $x_0$ with $y_0$, $x_2$ with $y_2$, and so forth, while keeping the other layers unchanged.

\subsubsection{Feature Map Channel-Exchange (CE)}
Feature map channel exchange uses fixed rules for exchanging channels, as shown in Figure~\ref{fig:seed_exchange} (b). For instance, in channel exchange, parts of the channels are selected based on a predefined step length (e.g., $2$) for the exchange. Assuming $x$ channels, we exchange channels $0,2,4,\ldots$ with the corresponding channels from another image, and so on.

\begin{figure*}[!t]
  \centering
  \includegraphics[width=\textwidth]{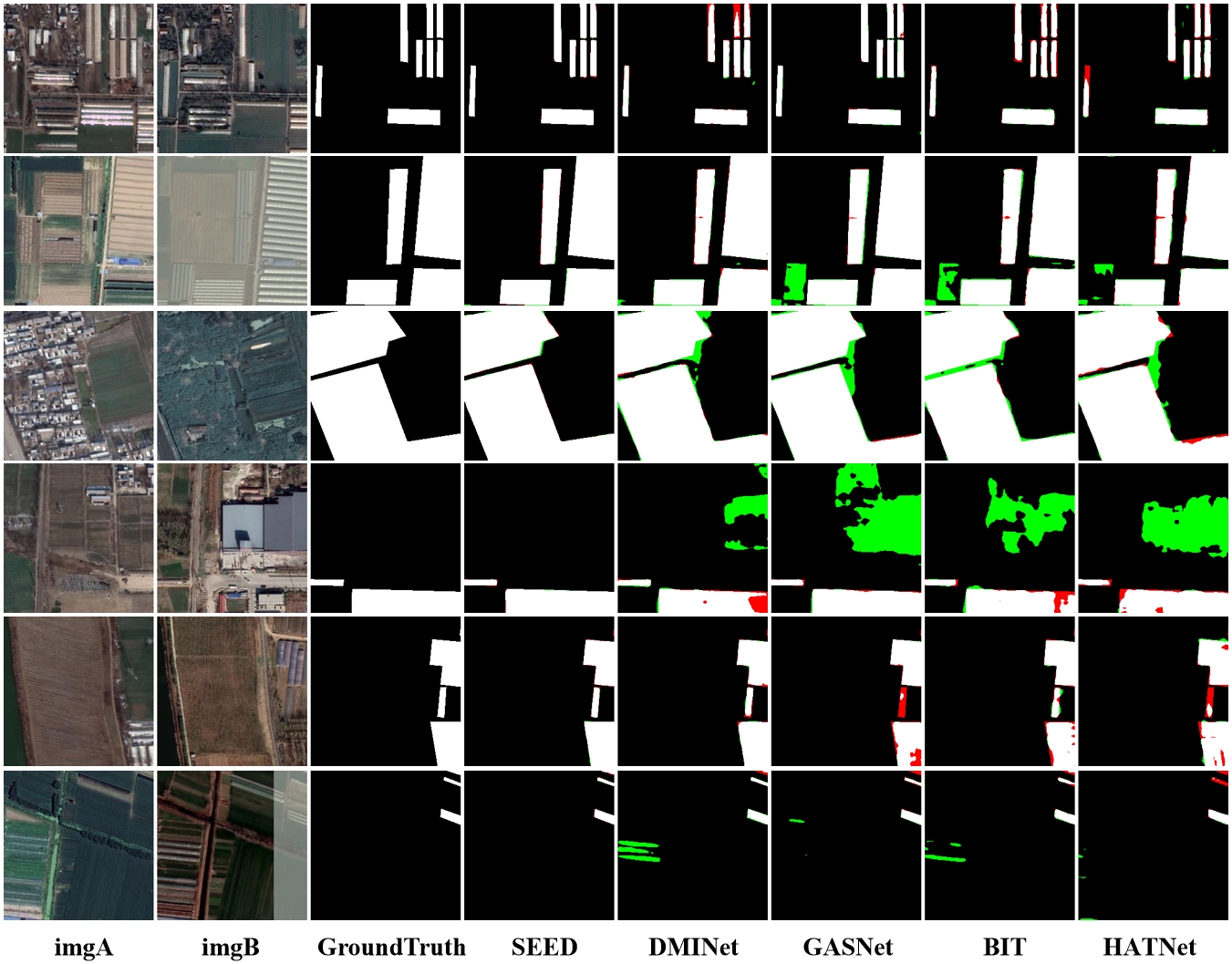}
  \caption{Visualization results in PX-CLCD dataset.}
  \label{fig:seed_pxclcd}
\end{figure*}

\subsubsection{Feature Map Spatial-Exchange (SE)}
Feature map spatial exchange performs feature exchange in the spatial dimension, as illustrated in Figure~\ref{fig:seed_exchange} (c). Specifically, given two feature maps from bi-temporal images, a predefined step length (e.g., $2$) is used to partition the feature maps over the width (or height) dimension at the specified intervals and swap the corresponding columns (or rows) between the two feature maps.

\begin{table}[htbp]
\centering
\caption{Quantitative Results On The PX-CLCD Dataset}
\label{tab:px_clcd_results}
\begin{tabular}{lccccc}
\hline
\textbf{Model} & \textbf{OA} & \textbf{IoU} & \textbf{F1} & \textbf{Rec} & \textbf{Prec} \\
\hline
HANet \cite{han_hanet_nodate}         & 98.50 & 88.99 & 94.18 & 93.83 & 94.53 \\
MSCANet \cite{m_liu_cnn-transformer_2022}       & 98.50 & 89.00 & 94.18 & 93.95 & 94.41 \\
DDAM-Net \cite{feng_ddam-net_2024} & -- & 90.70 & 95.12 & 94.78 & 95.47 \\
BIT \cite{chen_remote_2022}           & 98.76 & 90.78 & 95.17 & 94.80 & 95.54 \\
GASNet \cite{zhang_global-aware_2023}    & 98.99 & 92.51 & 96.11 & 96.42 & 95.80 \\
DMINet \cite{feng_change_nodate}         & 99.04 & 92.83 & 96.28 & 96.31 & 96.25 \\
SNUNet3+ \cite{miao_snunet3_2024}     & 99.19 & 93.61 & 96.64 & 96.79 & 96.60 \\
CGNet \cite{han_change_2023}         & 99.17 & 93.82 & 96.81 & 97.33 & 96.30 \\
\hline
SEED (LE)          & 99.38 & 95.34 & 97.61 & 97.46 & 97.76 \\
SEED (CE)          & 99.40	& 95.50	& 97.70	& 98.07	& 97.33 \\
SEED (SE)          & 99.32 & 94.86 & 97.36 & 96.89 & 97.84 \\
\hline
\end{tabular}
\end{table}


\subsubsection{Randomized Exchange (RE)}
To enrich cross-temporal mixing during training and to further probe the pixel-consistency principle, we introduce stochastic variants of the three exchange schemes. At each feature level, a binary mask is sampled and used to construct a \emph{permutation operator} that swaps the selected components across time. For any sampled mask, the induced exchange matrix is a permutation (orthogonal) operator, hence information-preserving and invertible.

\paragraph{Random Layer Exchange (RLE)}
Let $\epsilon^\ell \sim \mathrm{Bernoulli}(p_L)$ be an i.i.d. indicator for each level $\ell\!\in\!\{1,\ldots,L\}$. If $\epsilon^\ell=1$, we swap $(X_A^\ell,X_B^\ell)\!\leftarrow\!(X_B^\ell,X_A^\ell)$; otherwise we keep them unchanged.

\paragraph{Random Channel Exchange (RCE)}
For $X_A^\ell,X_B^\ell\in\mathbb{R}^{C\times H\times W}$, we sample a channel mask
$m^\ell\in\{0,1\}^{C}$ with $m^\ell_c\sim\mathrm{Bernoulli}(p_C)$ and broadcast it to
$\tilde{m}^\ell\in\{0,1\}^{C\times H\times W}$.
We then compute
\begin{align}
\bar{X}_A^\ell &= \tilde{m}^\ell\odot X_B^\ell + (1-\tilde{m}^\ell)\odot X_A^\ell,\\
\bar{X}_B^\ell &= \tilde{m}^\ell\odot X_A^\ell + (1-\tilde{m}^\ell)\odot X_B^\ell,
\end{align}
and set $X_A^\ell\leftarrow \bar{X}_A^\ell,\; X_B^\ell\leftarrow \bar{X}_B^\ell$.

This swaps the channels selected by $m^\ell$ while leaving others intact.

\paragraph{Random Spatial Exchange (RSE)}
We perform stochastic exchange along one spatial dimension (column-wise by default).
For $X_A^\ell, X_B^\ell \in \mathbb{R}^{C\times H\times W}$, we sample a binary column mask
$s^\ell \in \{0,1\}^{W}$ with $s^\ell_j \sim \mathrm{Bernoulli}(p_S)$.
We broadcast it to $\tilde{s}^\ell \in \{0,1\}^{C\times H\times W}$ by
$\tilde{s}^\ell_{c,h,j} = s^\ell_j$ for all $c\in\{1,\ldots,C\}$ and $h\in\{1,\ldots,H\}$.
Then the exchanged features are computed by
\begin{align}
\bar{X}_A^\ell &= \tilde{s}^\ell \odot X_B^\ell + \big(1-\tilde{s}^\ell\big)\odot X_A^\ell, \\
\bar{X}_B^\ell &= \tilde{s}^\ell \odot X_A^\ell + \big(1-\tilde{s}^\ell\big)\odot X_B^\ell,
\end{align}
and we set $X_A^\ell \leftarrow \bar{X}_A^\ell,\; X_B^\ell \leftarrow \bar{X}_B^\ell$.
A row-wise variant is analogous by sampling $r^\ell \in \{0,1\}^{H}$ and broadcasting along the height dimension.

\paragraph{Training and inference}
During training, masks are re-sampled at every iteration (per mini-batch) with user-chosen probabilities $p_L,p_C,p_S$ (default $0.5$). During validation and testing, we disable randomness and use a fixed deterministic exchange (LE/CE/SE as specified for the experiment). Randomized exchange adds negligible computation and acts as a stochastic, information-preserving permutation regularizer.

\begin{figure*}[!t]
  \centering
  \includegraphics[width=\textwidth]{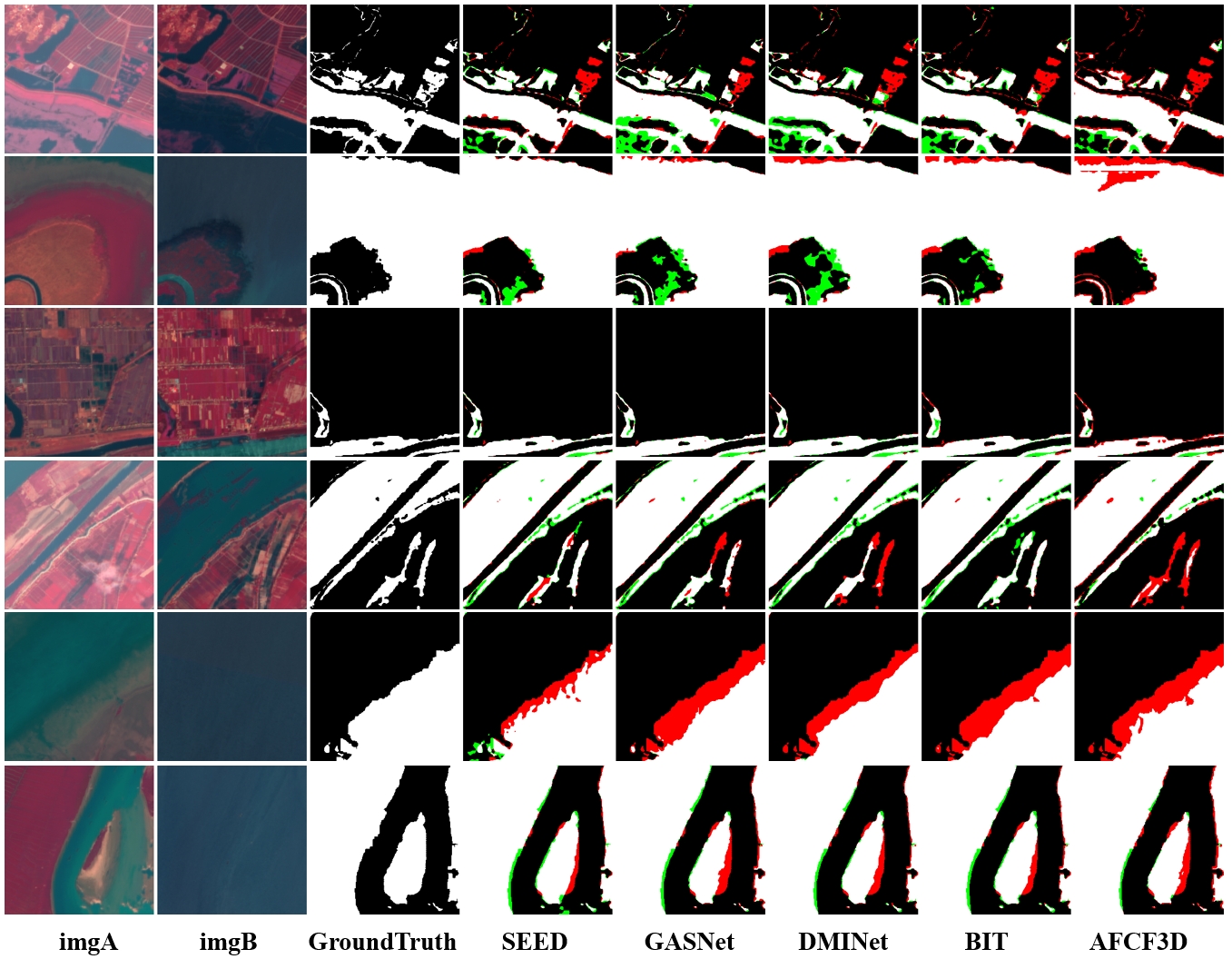}
  \caption{Visualization results in WaterCD dataset.}
  \label{fig:seed_watercd}
\end{figure*}

We conducted extensive experiments with the three exchange schemes (LE/CE/SE) and their randomized variants. Across backbones and datasets, we observed no statistically significant degradation and, in several settings, small but consistent improvements in IoU/F1. For randomized exchange, although prior work in vision often associates channels with semantics and spatial layouts with texture, our results show that change detection remains stable when swapping selected channels or spatial locations---even under stochastic policies.

These findings are consistent with our theory: each exchange is a permutation (orthogonal) operator that preserves information, and the pixel-consistency principle ensures that the learning objective (deciding whether corresponding pixels change) is unchanged by such permutations. Consequently, deep models need not rely on explicit difference construction; they can learn change cues directly from bi-temporal features under exchange. Unless otherwise stated, we apply randomized exchange during training and disable randomness at validation/test time by using a fixed deterministic exchange (LE/CE/SE).

\begin{table}[htbp]
\centering
\caption{Quantitative Results On The WaterCD Dataset}
\label{tab:water_cd_results}
\begin{tabular}{lccccc}
\hline
\textbf{Model} & \textbf{OA} & \textbf{IoU} & \textbf{F1} & \textbf{Rec} & \textbf{Prec} \\
\hline
AFCF3DNet \cite{ye_adjacent-level_2023} & 95.63 & 77.06 & 87.04 & 80.17 & 95.20 \\
HANet \cite{han_hanet_nodate}    & 95.94 & 79.94 & 88.85 & 88.34 & 89.37 \\
ELGCNet \cite{m_noman_elgc-net_2024}   & 96.14 & 80.79 & 89.37 & 88.60 & 90.16 \\
LU-Net \cite{s_sun_l-unet_2022}        & 96.44 & 82.10 & 90.17 & 89.07 & 91.30 \\
BIT \cite{chen_remote_2022}           & 96.52 & 82.52 & 90.42 & 89.55 & 91.31 \\
DMINet \cite{feng_change_nodate}         & 96.55 & 82.67 & 90.51 & 89.87 & 91.17 \\
GASNet \cite{zhang_global-aware_2023}    & 96.72 & 83.48 & 91.00 & 90.41 & 91.59 \\
CDNeXt \cite{wei_robust_2024}        & 96.88 & 84.21 & 91.43 & 90.71 & 92.15 \\
TRSANet \cite{j_li_trsanet_2024}     & 97.01 & 84.69 & 91.71 & 90.22 & 93.25 \\
\hline
SEED (LE)          & 96.98 & 84.64 & 91.68 & 90.69 & 92.69 \\
SEED (CE) & 96.94	& 84.43	& 91.56	& 90.64	& 92.50 \\
SEED (SE) & 96.86	& 84.06	& 91.34	& 90.42	& 92.27 \\
\hline
\end{tabular}
\end{table}

\subsection{Formalization and Invertibility Proof of Feature Exchange Strategy}
In the previous sections, different feature exchange strategies have been described intuitively, but no strict mathematical definitions were provided. To further demonstrate the effectiveness and information-preserving nature of the feature exchange strategy, this section provides a formal description of feature exchange strategies using random feature exchange as an example, and presents the proof of invertibility.

\subsubsection{Bernoulli Modeling and Matrix Formulation of Random Exchange}
Let the two feature vectors be $\boldsymbol{x}, \boldsymbol{y} \in \mathbb{R}^{m}$, and let the merged vector be $\boldsymbol{z} = \big[ \boldsymbol{x}^{\top},\, \boldsymbol{y}^{\top}\big]^{\top} \in \mathbb{R}^{2m}$.

For each index $i \in \{1,\ldots,m\}$ (representing layer/channel/spatial position), perform Bernoulli sampling:
\begin{align}
\epsilon_i \overset{\text{i.i.d.}}{\sim} \mathrm{Bernoulli}(p) \\
\mathbb{P}(\epsilon_i=1)=p \\
\mathbb{P}(\epsilon_i=0)=1-p
\end{align}
where $\epsilon_i=1$ means "swap this position", and $\epsilon_i=0$ means "do not swap". The exchange matrix can be constructed as:
\begin{align}
D=\mathrm{diag}(\epsilon_1,\ldots,\epsilon_m)\\
E=\Id_m-D
\end{align}
Next, define the general exchange operator as follows:
\begin{align}
P_{\mathrm{swap}} &=
\begin{bmatrix} E & D \\ D & E \end{bmatrix}
\in \mathbb{R}^{2m\times 2m}
\end{align}
The matrix computation for feature exchange can then be written as:
\begin{align}
\begin{bmatrix} \boldsymbol{x}'\\  \boldsymbol{y}'\end{bmatrix}
&=
P_{\mathrm{swap}}
\begin{bmatrix} \boldsymbol{x}\\  \boldsymbol{y}\end{bmatrix}
=
\begin{bmatrix}
E \boldsymbol{x}+D \boldsymbol{y}\\
D \boldsymbol{x}+E \boldsymbol{y}
\end{bmatrix}
\end{align}
For the $i$-th component, let $E_{ii}=1-\epsilon_i,\ D_{ii}=\epsilon_i$, then we obtain:
\begin{align}
x'_i=(1-\epsilon_i)x_i+\epsilon_i y_i \\
y'_i=\epsilon_i x_i+(1-\epsilon_i)y_i
\end{align}
Thus, the following can be computed:
\begin{align}
\epsilon_i=0\ \Rightarrow\ (x'_i,y'_i)=(x_i,y_i)\ \text{(no swap)}\\
\epsilon_i=1\ \Rightarrow\ (x'_i,y'_i)=(y_i,x_i)\ \text{(swap)}
\end{align}
Therefore, through the matrix computation, the "switchable exchange" for any index $i$ is achieved.

\subsubsection{Orthogonal Invertibility Proof of Feature Exchange Matrix}
Since $\epsilon_i \in \{0, 1\}$, we have $D^2 = D$; let $E = \Id_m - D$, then
$E^2 = E,\ DE = ED = 0,\ D+E = \Id_m$. This gives:
\begin{align}
P_{\mathrm{swap}}^{\top}P_{\mathrm{swap}}
&=
\begin{bmatrix}
E^2+D^2 & ED+DE\\
DE+ED & D^2+E^2
\end{bmatrix} \notag\\
&=
\begin{bmatrix}
\Id_m & 0\\
0 & \Id_m
\end{bmatrix} \notag\\
&= \Id_{2m}
\end{align}
Thus, $P_{\mathrm{swap}}$ is a permutation matrix, so it is orthogonal and invertible: $P_{\mathrm{swap}}^{-1} = P_{\mathrm{swap}}^{\top}$, and $\det(P_{\mathrm{swap}}) = \pm 1$.

\subsubsection{Relationship Between Random Exchange Sampling and Invertibility}
The orthogonal invertibility conclusion holds for each specific exchange result per sample: for any binary mask $\epsilon \in \{0,1\}^m$, let $S = \{i : \epsilon_i = 1\}$, $D = \mathrm{diag}(\mathbf{1}\{i \in S\})$, $E = \Id_m - D$, then
\begin{align}
P_{\mathrm{swap}} = \begin{bmatrix} E & D \\ D & E \end{bmatrix} \\
P_{\mathrm{swap}}^{\top}P_{\mathrm{swap}} = \Id_{2m} \\
\det(P_{\mathrm{swap}}) = (-1)^{|S|}
\end{align}
Thus, invertibility/orthogonality is independent of how $S$ is sampled: whether it is independent Bernoulli, fixed $k$-selection, group selection, or data-dependent, as long as $\epsilon_i \in \{0,1\}$, the above conclusions hold.

\subsubsection{Feature Exchange Superiority Analysis from Information Theoretic and Geometric Perspectives}
Let $ \boldsymbol{Z} = [ \boldsymbol{x}^{\top}, \boldsymbol{y}^{\top}]^{\top} \in \mathbb{R}^{2m}$, with label $Y$. If the exchange is performed by the orthogonal permutation matrix $P_{\mathrm{swap}} \in \mathbb{R}^{2m \times 2m}$ (as proven earlier that $P_{\mathrm{swap}}^{\top} P_{\mathrm{swap}} = \Id_{2m}$), then:
\begin{align}
\MI(P_{\mathrm{swap}} \boldsymbol{Z}; Y) = \MI(\boldsymbol{Z}; Y) \\
\mathcal{R}^\star(P_{\mathrm{swap}} \boldsymbol{Z} \to Y) = \mathcal{R}^\star(\boldsymbol{Z} \to Y)
\end{align}
where:
\begin{itemize}
  \item Mutual Information $\MI(U; Y)$ measures the amount of information and discriminative power about the label $Y$ contained in the feature $U$. A higher value indicates that the processed feature better explains or predicts $Y$ (an invertible transformation like orthogonal permutation does not change this, so no information is lost). Bayes' minimum risk $\mathcal{R}^\star(U \to Y)$ represents the theoretical minimum error rate or optimal performance upper bound achievable using feature $U$ under a given loss. If this value remains unchanged after processing, it indicates that the optimal performance achievable in the task is not affected, i.e., it does not degrade change detection accuracy.
  \item $\MI(P_{\mathrm{swap}} \boldsymbol{Z}; Y) = \MI(\boldsymbol{Z}; Y)$: Orthogonal permutation is an invertible isometric transformation that neither adds nor loses information about the label $Y$; mutual information remains unchanged (data processing inequalities hold with equality under invertible mappings).
  \item $\mathcal{R}^\star(\cdot \to Y)$: This represents the Bayes minimum risk under a given loss. Equality on both sides indicates that after the permutation, the theoretically optimal recognition performance remains unchanged (an optimal decision function exists in a one-to-one correspondence).
\end{itemize}

In contrast, common "differential feature" methods are usually non-invertible or non-orthogonal, and thus typically reduce mutual information or worsen numerical conditions. For example, simply adding or subtracting, i.e., $U = \boldsymbol{x} + \boldsymbol{y}$ or $V = \boldsymbol{x} - \boldsymbol{y}$, corresponds to linear mappings $S_U = [\Id_m\ \ \Id_m]$ or $S_V = [\Id_m\ \ -\Id_m]$, both with rank $m < 2m$, which are non-invertible. By the data processing inequality:
\begin{align}
\MI(U; Y) \le \MI(\boldsymbol{Z}; Y)\\
\MI(V; Y) \le \MI(\boldsymbol{Z}; Y)
\end{align}
Thus, compared to feature exchange, operations like addition or subtraction are information-losing processes.

If concatenation + $1\times1$ linear dimensionality reduction is used, then $\tilde{\boldsymbol{Z}} = [\boldsymbol{x}; \boldsymbol{y}]$ and a $W \in \mathbb{R}^{C' \times 2C}$ is applied. If $C' < 2C$ (commonly seen with compressed channel numbers), $\operatorname{rank}(W) \le C' < 2C$, which is non-invertible, and thus:
\begin{align}
\MI(W\tilde{\boldsymbol{Z}}; Y) \le \MI(\tilde{\boldsymbol{Z}}; Y) = \MI(\boldsymbol{Z}; Y)
\end{align}

In conclusion, this paper formalizes feature exchange as a permutation operator generated by a binary mask $P_{\mathrm{swap}} = \begin{bmatrix} E & D \\ D & E \end{bmatrix}$ (where $D = \mathrm{diag}(\epsilon)$, $E = \Id - D$), and rigorously proves that $P_{\mathrm{swap}}^{\top} P_{\mathrm{swap}} = \Id$, meaning that $P_{\mathrm{swap}}$ is a permutation matrix, orthogonal, and invertible. This leads to two direct conclusions: first, mutual information remains unchanged:
\begin{align}
\MI(P_{\mathrm{swap}} \boldsymbol{Z}; Y) = \MI(\boldsymbol{Z}; Y)
\end{align}
second, Bayes' minimum risk remains unchanged:
\begin{align}
\mathcal{R}^\star(P_{\mathrm{swap}} \boldsymbol{Z} \to Y) = \mathcal{R}^\star(\boldsymbol{Z} \to Y)
\end{align}
Therefore, feature exchange is an information-preserving, performance-preserving preprocessing technique, and due to its orthogonality, it is more numerically advantageous for optimization. In contrast, standalone addition/subtraction (dimensionality reduction) and "concatenation + $1 \times 1$ linear dimensionality reduction" are typically non-invertible or non-orthogonal, which in most cases reduces mutual information or worsens numerical conditions. Hence, from both an information-theoretic and geometric perspective, permutation-based feature exchange is a safer, more robust design choice and provides theoretical support for the excellent performance of the SEED framework without introducing additional learnable difference modules.

\subsection{Siamese Encoder-Exchange-Decoder}
Feature exchange can perturb representations in semantic segmentation. In change detection, the goal is to decide whether corresponding pixels across time have changed. As long as spatial correspondence is preserved, this goal does not change. We refer to this requirement as the \emph{pixel-consistency principle}. Under pixel consistency, zero-parameter exchanges can be applied without altering the supervision target and typically without degrading performance.

Given bi-temporal inputs $(I_A, I_B)$, a weight-sharing Siamese encoder $E_\psi$ produces two feature pyramids $\{X_A^\ell\}_{\ell=1}^L$ and $\{X_B^\ell\}_{\ell=1}^L$. An exchange stage then applies a parameter-free permutation operator $\Pi^\ell$ at each level $\ell$, yielding exchanged features
\begin{align}
  \tilde{X}_A^\ell,\,\tilde{X}_B^\ell \;=\; \Pi^\ell\!\big(X_A^\ell, X_B^\ell\big), \quad \ell = 1,\dots,L.
\end{align}

A Feature Pyramid Network (FPN)~\cite{lin_feature_2017} with shared parameters $F_\theta$ is applied \emph{separately} to each branch (no cross-branch concatenation/addition), producing
\begin{align}
Y_A^\ell \;=\; F_\theta(\tilde{X}_A^\ell) \\
Y_B^\ell \;=\; F_\theta(\tilde{X}_B^\ell)
\end{align}
A lightweight layer-by-layer upsampling decoder with shared weights $D_\phi$ maps each pyramid to logits:
\begin{align}
Z_A \;=\; D_\phi\big(\{Y_A^\ell\}_{\ell=1}^L\big) \\
Z_B \;=\; D_\phi\big(\{Y_B^\ell\}_{\ell=1}^L\big)
\end{align}

During training, both branches predict a binary change map and the total loss is the sum over branches. At inference, we average branch logits before the final sigmoid:
\begin{align}
\hat{Y} \;=\; \sigma\!\left(\tfrac{1}{2}\big(Z_A + Z_B\big)\right)
\end{align}
Because the exchange is parameter-free and FPN/decoder weights are shared across branches, SEED effectively uses a single encoder-decoder parameter set while avoiding any explicit fusion layer between times.


\begin{figure*}[!t]
  \centering
  \includegraphics[width=\textwidth]{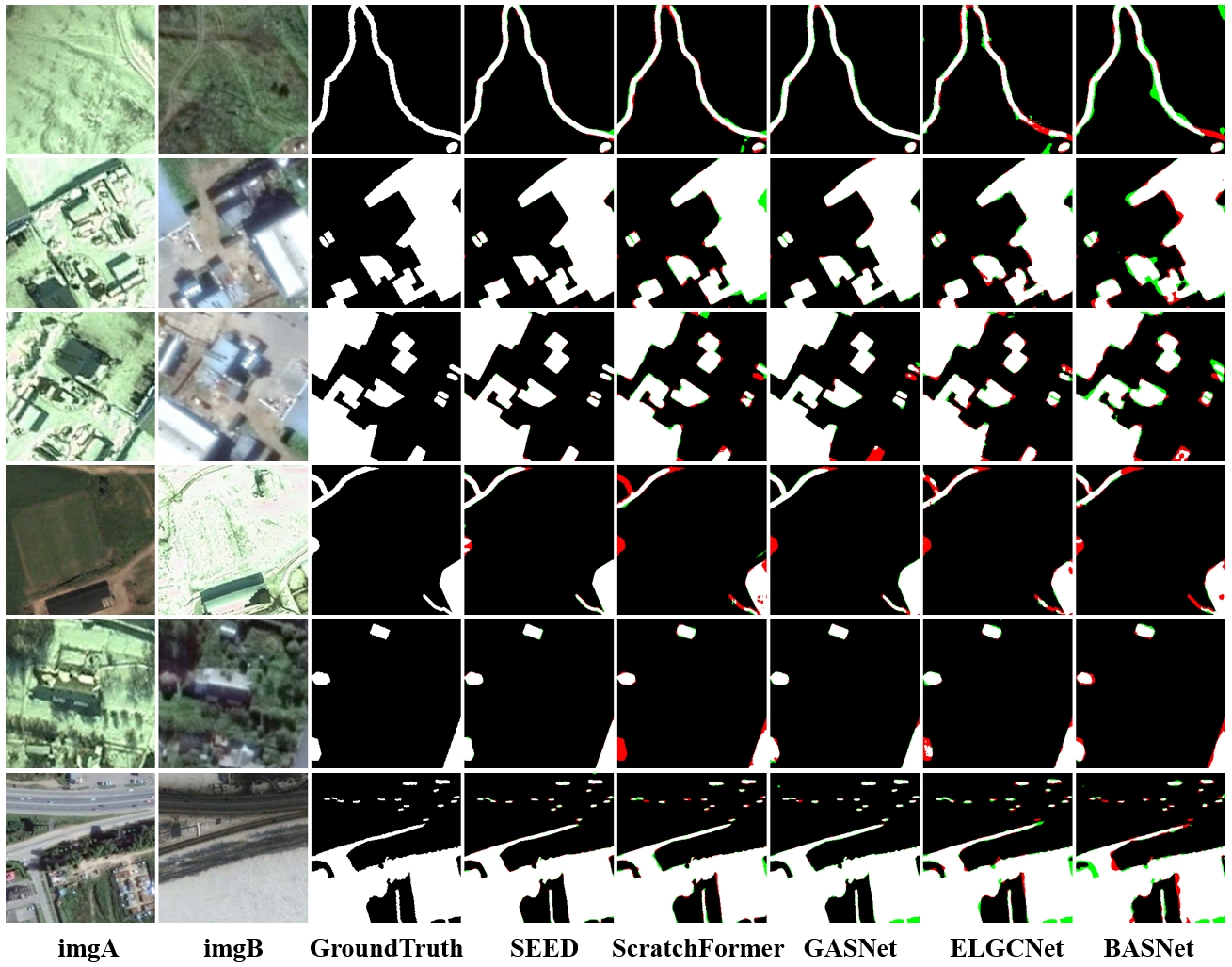}
  \caption{Visualization results in CDD dataset.}
  \label{fig:seed_cdd}
\end{figure*}


\begin{table}[htbp]
\centering
\caption{Quantitative Results On The CDD Dataset}
\label{tab:cdd_results}
\begin{tabular}{lccccc}
\hline
\textbf{Model} & \textbf{OA} & \textbf{IoU} & \textbf{F1} & \textbf{Rec} & \textbf{Prec} \\
\hline
BASNet \cite{z_wang_bitemporal_2024} & 99.18 & 93.29 & 96.53 & 96.30 & 96.76 \\  
UA-BCD \cite{li_overcoming_2025}     & --    & 93.49 & 96.64 & 96.90 & 96.38 \\
RCDT \cite{lu_cross_2024}                  & --    & 93.79 & 96.80 & 96.97 & 96.63 \\
CDMamba \cite{zhang_cdmamba_2025}    & 99.26 & 93.93 & 96.87 & 96.84 & 96.90 \\
SFFCE-CD \cite{y_xing_sffce-cd_2025}   & 99.29 & 94.42 & 97.13 & 96.94 &  97.32 \\
ELGCNet \cite{m_noman_elgc-net_2024}   & 99.33  &  94.50   & 91.17  & --   & -- \\   
DgFA \cite{f_zhou_dual-granularity_2025} & 99.41 & 95.12 & 97.50 & 97.60 & 97.40 \\
GASNet \cite{zhang_global-aware_2023}  & 99.41  & 95.34  & 97.61  & 98.06  & 97.17 \\   
MLDFNet \cite{d_sidekejiang_mldfnet_2025}  & -- & 95.78 & 97.84 & 97.97 & 97.72 \\
ScratchFormer \cite{m_noman_remote_2024}   & 99.50 & 95.85  & 97.88 & -- & -- \\   
FTransDF-Net \cite{li_dual_2025}   & -- & 95.85 &  97.88 & 97.63 & 98.13 \\
DSFI-CD \cite{x_li_dsfi-cd_2025}  & 99.52 & 96.10 &  98.01 & 98.34 & 97.68 \\
HASNet \cite{c_tao_hasnet_2025}        & 99.56 & 96.49 & 98.21 & 98.12 & 98.32 \\
\hline
SEED (LE)                               & 99.64	& 97.11	& 98.53	& 98.44	& 98.63 \\
SEED (CE) & 99.59	& 96.75	& 98.35	& 98.36	& 98.34 \\
SEED (SE) & 99.55	& 96.40	& 98.17	& 98.02	& 98.32 \\
\hline
\end{tabular}
\end{table}

\section{Experiments}
\subsection{Datasets}
In this study, we utilized five sub-meter resolution datasets—SYSU-CD~\cite{shi_deeply_2022}, LEVIR-CD~\cite{chen_spatial-temporal_2020}, PX-CLCD~\cite{miao_snunet3_2024}, WaterCD~\cite{j_li_trsanet_2024}, and CDD~\cite{lebedev_change_2018}—to demonstrate the robustness and versatility of our change detection algorithm across different environments and scenarios. The SYSU-CD dataset comprises a total of 20,000 images with dimensions of 256×256, in which changes of multiple targets are annotated. The dataset is divided into three sets: the training set, validation set, and test set, consisting of 12,000, 4,000, and 4,000 image pairs, respectively.The LEVIR-CD dataset is an important resource for building change detection. It is divided into 445 training image pairs, 64 validation image pairs, and 128 testing image pairs. During training, we perform random cropping to 256×256 patches, while during testing, a sliding window approach is adopted to obtain the results.The PX-CLCD dataset mainly targets cultivated land change detection and contains a total of 5,170 images with dimensions of 256×256. The dataset is split into training, validation, and test sets in a 6:2:2 ratio. The WaterCD dataset is a change detection dataset focused on water resources, with image dimensions of 512×512. The training, validation, and test sets contain 765, 207, and 177 images, respectively. During training, random 256×256 patches are used, and during inference, a sliding window approach is employed.The CDD dataset is designed for detecting seasonal changes, with image dimensions of 256×256. The training, validation, and test sets contain 10,000, 3,000, and 3,000 images, respectively.

\subsection{Implementation Details}
We trained the SEED on the NVIDIA A100 and V100 GPUs. And, we used three methods: RandomRotate, RandomFlip and PhotoMetricDistortion for data enhancement. In terms of model optimization, the AdamW optimizer was utilized. Throughout the experimental stage, we continuously monitored the IoU metric on the validation set, earmarking the best-performing model for subsequent final evaluation. In the experiments of this paper, we keep the IoU metric as the primary evaluation criterion.

\subsection{Evaluation Metrics}
We used these metrics to evaluate our model like precision (Prec), recall (Rec), overall accuracy (OA), F1-score (F1), and Intersection over Union (IoU). Its calculation formula is as follows:
\begin{align}
\mathrm{IoU} = \frac{TP}{TP + FN + FP} \\
\mathrm{Prec} = \frac{TP}{TP + FP} \\
\mathrm{Rec} = \frac{TP}{TP + FN} \\
F1 = \frac{2 \cdot Prec \cdot Rec}{Prec + Rec} \\
\mathrm{OA} = \frac{TP + TN}{TP + TN + FN + FP}
\end{align}

\subsection{Quantitative analysis and visualize results with compared methods}
In this article, we conducted detailed comparative experiments between the SEED architecture and the state-of-the-art change detection algorithms. All the SEED's results in these tables depended on the Swin Transformer V2 backbone. As shown in Table~\ref{tab:sysu_cd_results} to Table~\ref{tab:cdd_results}, the SEED framework consistently performs outstandingly across different datasets, fully demonstrating its effectiveness and robustness. In addition, whether it is layer exchange (LE), channel exchange (CE), or spatial exchange (SE), the accuracy metrics of the SEED architecture are outstanding. This further proves the rationality of feature exchange.

\begin{figure*}[!t]
    \centering
    \includegraphics[width=\textwidth]{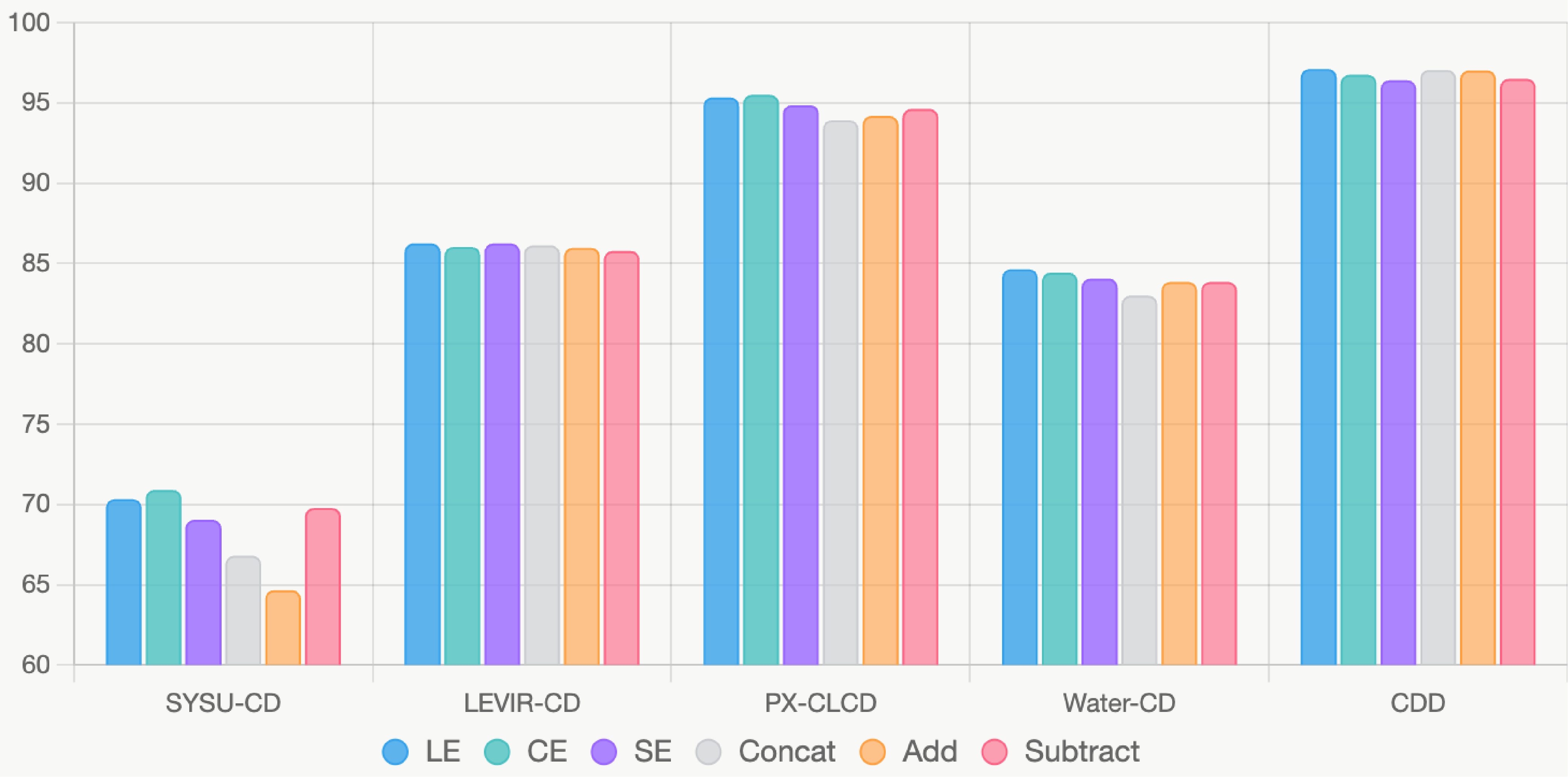} %
    \caption{Comparison results of different exchange methods based on IoU indicator on change detection datasets.}
    \label{fig:seed_compare_vis}
\end{figure*}

First, on the SYSU-CD dataset, the SEED model achieved an overall accuracy (OA) of 92.16\% and an IoU of 70.91\%, with an F1 score of 82.98\%. This not only significantly surpasses models such as DSAMNet, P2V, DARNet, and STDF-CD, but also exhibits a balanced performance in both Recall and Precision. This indicates that, even in complex scenarios, SEED can effectively capture subtle changes in bi-temporal images.

Second, on the LEVIR-CD dataset, which is focused on building change detection and where most models already achieve high scores, SEED still ranks first with an OA of 99.26\% and an IoU of 86.25\%. Compared to the RSBuilding model—trained on a larger building dataset—SEED shows a small yet stable advantage in F1 and Precision, demonstrating its superior ability to accurately identify change regions in building detection tasks.

On the PX-CLCD dataset, where the change detection task mainly concerns cultivated land and requires the extraction of fine details, the SEED model achieved an OA of 99.40\% and an IoU as high as 95.50\%, with an F1 score of 97.70\%, far outperforming other models such as SNUNet3+ and CGNet. This result proves that the SEED framework possesses stronger detail extraction and discrimination capabilities, making it better suited for complex cultivated land change detection scenarios.

\begin{figure*}[!t]
    \centering
    \includegraphics[width=\textwidth]{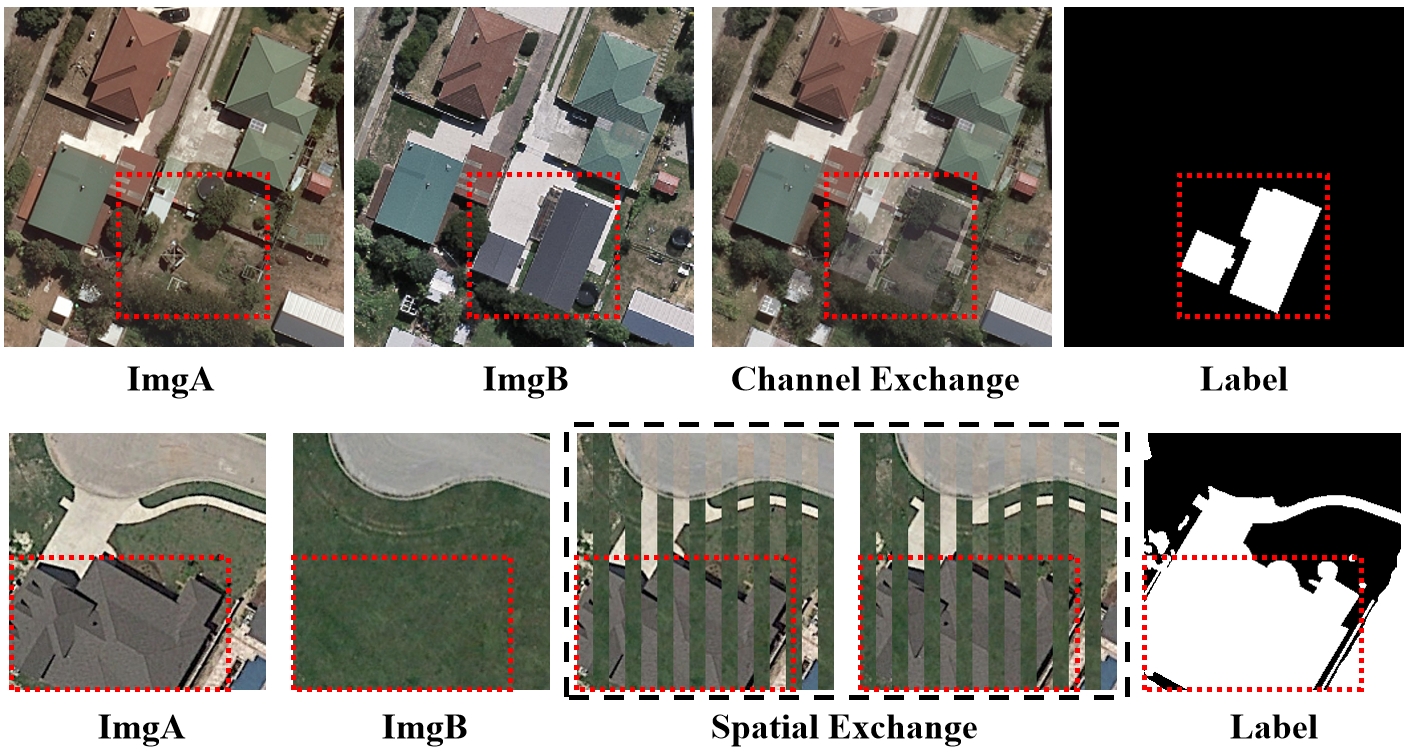} %
    \caption{The visualization of the exchange highlights the areas of change.}
    \label{fig:seed_exchange_vis}
\end{figure*}

\begin{figure*}[!t]
  \centering
    \includegraphics[width=\textwidth]{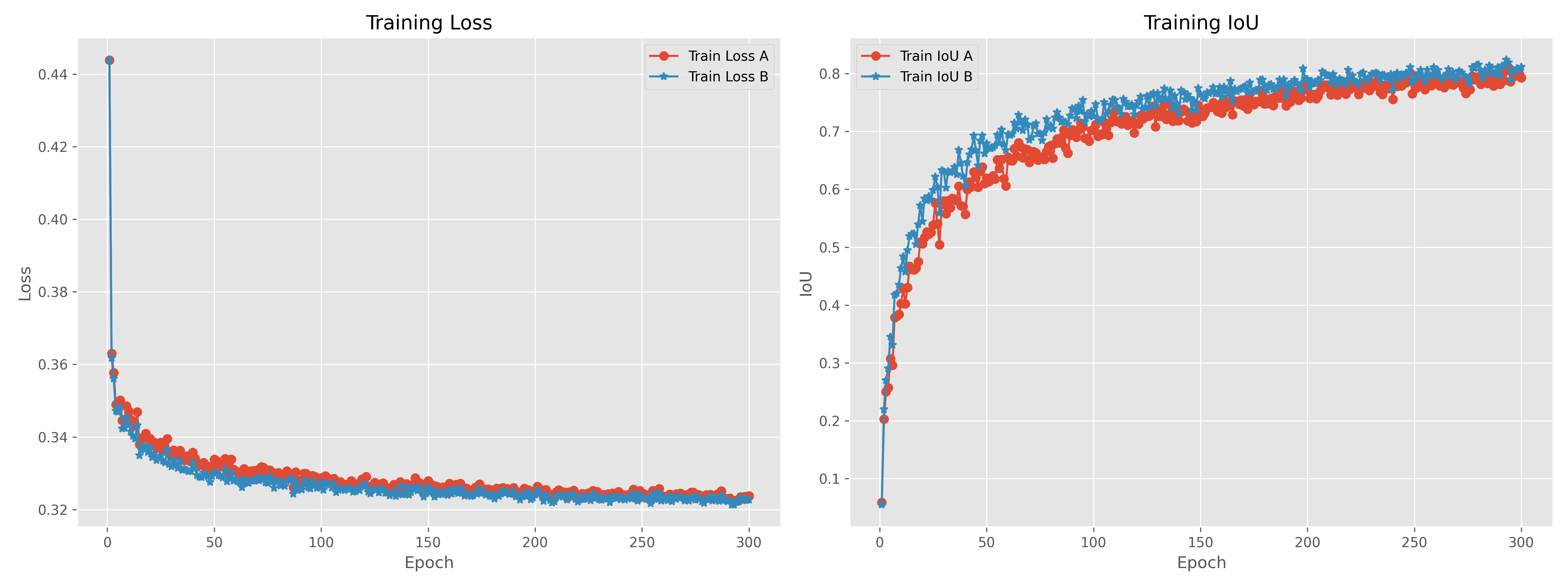} %
    \caption{Training log based on channel exchange.}
    \label{fig:channel_training}
\end{figure*}
  
\begin{figure*}[!t]
    \centering
    \includegraphics[width=\textwidth]{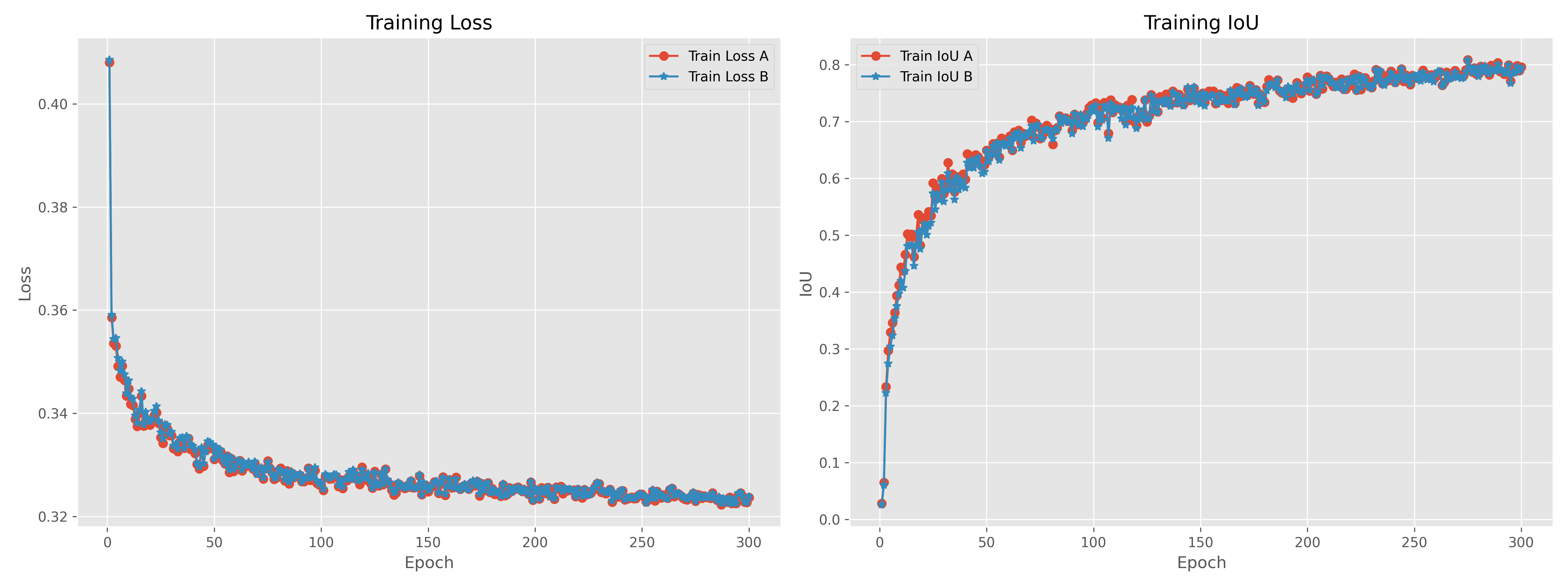} %
    \caption{Training log based on spatial exchange.}
    \label{fig:spatial_training}
\end{figure*}

For the WaterCD dataset, the SEED model still achieved an OA of 96.98\% and an IoU of 84.64\%, with an F1 score of 91.68\%, outperforming methods like AFCF3DNet, HANet, and ELGCNet across all metrics. This demonstrates that the SEED framework can effectively handle the complex features inherent in water resource change detection, offering high practical value.

Finally, on the CDD dataset, which focuses on seasonal changes and requires high precision in capturing subtle and continuous variations, the SEED model performed exceptionally well—leading all compared models with an OA of 99.64\%, an IoU of 97.11\%, and an F1 score of 98.53\%, thereby reflecting extremely high detection accuracy and stability.

Overall, by abandoning traditional difference feature computation and instead combining feature exchange with a simple encoder-decoder structure, the SEED framework not only achieves excellent performance across various scenarios and datasets but also demonstrates strong adaptability and robustness in applications such as building, cultivated land, water resource, and seasonal change detection. These experimental results fully validate the effectiveness of the feature exchange strategy in change detection tasks.

In addition, Figure~\ref{fig:seed_sysu} through Figure~\ref{fig:seed_cdd} present visualizations of the test results of the SEED model on five different datasets (SYSU-CD, LEVIR-CD, PX-CLCD, WaterCD, and CDD). These datasets represent various change detection scenarios: SYSU-CD covers multi-target changes, LEVIR-CD mainly focuses on building changes, PX-CLCD emphasizes cultivated land changes, WaterCD concentrates on water body changes, and CDD reflects the characteristics of seasonal changes. In these images, true positives (TP) are depicted as white pixels, true negatives (TN) as black pixels, false positives (FP) as green pixels, and false negatives (FN) as red pixels. The intuitive visual results demonstrate that the SEED model can accurately capture change regions across diverse scenarios, with its detection outcomes closely aligning with the ground truth annotations. It not only exhibits extremely high detection accuracy but also shows outstanding robustness and stability. These rich visualizations fully validate the excellent performance of the SEED framework in handling various change types and complex environments.

\subsection{Ablation studies}
In this paper, we propose a novel change detection framework—the SEED architecture. This framework completely dispenses with the commonly used difference feature computation modules in change detection tasks, and instead applies feature exchange solely on the bi-temporal feature pyramid. As shown in Table~\ref{tab:sysu_cd_backbone} to Table~\ref{tab:cdd_backbone}, we conducted experiments on the SYSU-CD, LEVIR-CD, PX-CLCD, WaterCD, and CDD datasets. For feature extraction, we respectively employed Swin Transformer V2, EfficientNet-B4, and ResNet50 as the backbone networks. During the feature exchange stage, three methods were adopted: layer exchange, channel exchange, and spatial exchange. In the decoder design, for the Swin Transformer V2-based backbone, the decoder was constructed using a Swin Transformer Block, while for the EfficientNet-B4 and ResNet50 backbones, the BottleNeck module from ResNet was used for feature optimization. For comparison with traditional difference feature construction methods, we selected concatenation, addition, and subtraction approaches. The experimental results indicate that, under most conditions, the performance of the SEED framework combined with feature exchange is superior to that of change detection frameworks based on traditional difference feature construction methods. As shown in Figure~\ref{fig:seed_compare_vis}, we draw a bar chart of different experimental results based on the IoU indicator. The left‑colored indicators represent the experiments that employ feature exchange, while the right‑colored indicators correspond to those that rely on traditional difference‑feature learning. In the bar chart, the left‑colored bars consistently appear taller, indicating that the feature‑exchange–based change‑detection architecture delivers superior performance compared with the architecture built on conventional difference‑feature learning.

Furthermore, we calculated the parameter counts for the layer exchange, channel exchange, spatial exchange, concatenation, addition, and subtraction methods. As shown in Table~\ref{tab:params_gflops}, the parameter count required by the feature exchange methods is identical to that of the addition and subtraction methods, and slightly lower than that of the addition method. This confirms that the SEED framework maintains consistency in parameter configuration with traditional difference feature construction methods, and further demonstrates that, due to its simple architecture, the SEED framework can achieve outstanding performance in change detection tasks.

\begin{table*}[htbp]
\centering
\caption{Test Results of Different Backbones Combined With Various Feature Exchange Mechanisms On The SYSU-CD Dataset}
\label{tab:sysu_cd_backbone}
\begin{tabular}{l c c c c c c}
\hline
\textbf{Backbone} & \textbf{Exchange Method} & \textbf{OA} & \textbf{IoU} & \textbf{F1} & \textbf{Rec} & \textbf{Prec} \\
\hline
\multirow{6}{*}{SwinT V2}
 & LE    & 92.03 & 70.33 & 82.58 & 80.16 & 85.16 \\
 & CE  & 92.16 & 70.91	& 82.98	& 81.01	& 85.05 \\
 & SE  & 91.52 & 69.05 & 81.69 & 80.27 & 83.16 \\
\cline{2-7}
 & Concat           & 91.24 & 66.81 & 80.10 & 74.76 & 86.27 \\
 & Add              & 90.42 & 64.66 & 78.54 & 74.32 & 83.27 \\
 & Subtract            & 92.08 & 69.78 & 82.20 & 77.50 & 87.51 \\
\hline
\multirow{6}{*}{ResNet50}
 & LE    & 91.87 & 68.54 & 81.34 & 75.07 & 88.74 \\
 & CE  & 91.84 & 69.96 & 82.33 & 80.57 & 84.16 \\
 & SE  & 91.86 & 69.42 & 81.95 & 78.32 & 85.93 \\
\cline{2-7}
 & Concat           & 91.45 & 68.24 & 81.13 & 77.94 & 84.58 \\
 & Add              & 91.22 & 67.99 & 80.94 & 79.03 & 82.95 \\
 & Subtract            & 91.20 & 67.75 & 80.78 & 78.41 & 83.29 \\
\hline
\multirow{6}{*}{EfficientNet-B4}
 & LE    & 92.20 & 70.01 & 82.36 & 77.23 & 88.23 \\
 & CE  & 91.53 & 67.93 & 80.90 & 76.09 & 86.37 \\    
 & SE  & 91.60 & 68.75 & 81.48 & 78.34 & 84.88 \\    
\cline{2-7}
 & Concat            & 90.93	& 65.45	& 79.12	& 72.9	& 86.51 \\
 & Add               & 90.85	& 66.19	& 79.66	& 75.94	& 83.75 \\
 & Subtract             & 91.07	& 65.41	& 79.09	& 71.59	& 88.34 \\

\hline
\end{tabular}
\end{table*}

\begin{table*}[htbp]
\centering
\caption{Test Results Of Different Backbones Combined With Various Feature Exchange Mechanisms On The LEVIR-CD Dataset}
\label{tab:levir_cd_backbone}
\begin{tabular}{l c c c c c c}
\hline
\textbf{Backbone} & \textbf{Exchange Method} & \textbf{OA} & \textbf{IoU} & \textbf{F1} & \textbf{Rec} & \textbf{Prec} \\
\hline
\multirow{6}{*}{SwinT V2} 
 & LE    & 99.26 & 86.25 & 92.62 & 90.97 & 94.32 \\
 & CE  & 99.25 & 86.03 & 92.49 & 91.08 & 93.94 \\
 & SE  & 99.26 & 86.24 & 92.61 & 91.42 & 93.83 \\
\cline{2-7}
 & Concat            & 99.25 & 86.13 & 92.55 & 90.97 & 94.18 \\
 & Add               & 99.24 & 85.98 & 92.46 & 91.35 & 93.61 \\
 & Subtract             & 99.23 & 85.78 & 92.35 & 90.74 & 94.01 \\
\hline
\multirow{6}{*}{ResNet50} 
 & LE    & 99.21 & 85.26 & 92.04 & 89.63 & 94.59 \\ 
 & CE  & 99.21 & 85.22 & 92.02 & 89.83 & 94.32 \\
 & SE  & 99.22 & 85.52 & 92.19 & 90.55 & 93.89 \\    
\cline{2-7}
 & Concat            & 99.21 & 85.44 & 92.15 & 90.78 & 93.57 \\    
 & Add               & 99.14 & 84.38 & 91.53 & 90.85 & 92.22 \\
 & Subtract             & 99.19 & 84.94 & 91.85 & 90.12 & 93.66 \\
\hline
\multirow{6}{*}{EfficientNet-B4} 
 & LE    & 99.21 & 85.34  & 92.09 & 89.74 & 94.57 \\
 & CE  & 99.22 & 85.56	& 92.22	& 90.29	& 94.23 \\
 & SE  & 99.19 & 85.01	& 91.90	& 89.73	& 94.17 \\
\cline{2-7}
 & Concat            & 99.21	& 85.34	& 92.09	& 90.77	& 93.45 \\
 & Add               & 99.15	& 84.18	& 91.41	& 88.57	& 94.43 \\
 & Subtract             & 99.19	& 85.11	& 91.95	& 90.71	& 93.24 \\
\hline
\end{tabular}
\end{table*}

\begin{table*}[htbp]
\centering
\caption{Test Results Of Different Backbones Combined With Various Feature Exchange Mechanisms On The PX-CLCD Dataset}
\label{tab:pxclcd_backbone}
\begin{tabular}{l c c c c c c}
\hline
\textbf{Backbone} & \textbf{Exchange Method} & \textbf{OA} & \textbf{IoU} & \textbf{F1} & \textbf{Rec} & \textbf{Prec} \\
\hline
\multirow{6}{*}{SwinT V2} 
 & LE    & 99.38 & 95.34 & 97.61 & 97.46 & 97.76 \\
 & CE  & 99.40 & 95.50 & 97.70 & 98.07 & 97.33 \\
 & SE  & 99.32 & 94.86 & 97.36 & 96.89 & 97.84 \\
\cline{2-7}
 & Concat            & 99.19 & 93.91 & 96.86 & 96.56 & 97.15 \\
 & Add               & 99.23 & 94.18 & 97.00 & 96.93 & 97.07 \\
 & Subtract             & 99.29 & 94.62 & 97.24 & 96.75 & 97.72 \\
\hline
\multirow{6}{*}{ResNet50} 
 & LE    & 99.12 & 93.35 & 96.56 & 95.45 & 97.70 \\
 & CE  & 99.00 & 92.44 & 96.07 & 94.84 & 97.33 \\
 & SE  & 99.02 & 92.61 & 96.16 & 95.22 & 97.12 \\
\cline{2-7}
 & Concat            & 99.18 & 93.82 & 96.81 & 96.32 & 97.30 \\
 & Add               & 99.01 & 92.45 & 96.08 & 94.39 & 97.83 \\ 
 & Subtract             & 99.13 & 93.51 & 96.65 & 96.65 & 96.64 \\
\hline
\multirow{6}{*}{EfficientNet-B4} 
 & LE    & 99.13	& 93.41	& 96.59	& 95.76	& 97.44 \\
 & CE  & 99.00 & 92.49 & 96.10 & 95.02 & 97.20 \\
 & SE  & 98.98	& 92.32	& 96.01	& 95.23	& 96.79 \\
\cline{2-7}
 & Concat            & 99.09	& 93.19	& 96.47	& 96.72	& 96.23 \\
 & Add               & 98.85 & 91.42 & 95.52 & 95.05 & 95.99 \\
 & Subtract          & 99.10	& 93.23	& 96.50	& 96.26	& 96.73 \\

\hline
\end{tabular}
\end{table*}

\begin{table*}[htbp]
\centering
\caption{Test Results Of Different Backbones Combined With Various Feature Exchange Mechanisms On The WaterCD Dataset}
\label{tab:water_cd_backbone}
\begin{tabular}{l c c c c c c}
\hline
\textbf{Backbone} & \textbf{Exchange Method} & \textbf{OA} & \textbf{IoU} & \textbf{F1} & \textbf{Rec} & \textbf{Prec} \\
\hline
\multirow{6}{*}{SwinT V2} 
 & LE    & 96.98 & 84.64 & 91.68 & 90.69 & 92.69 \\
 & CE  & 96.94 & 84.43 & 91.56 & 90.64 & 92.50 \\
 & SE  & 96.86 & 84.06 & 91.34 & 90.42 & 92.27 \\
\cline{2-7}
 & Concat            & 96.64 & 83.00 & 90.71 & 89.55 & 91.91 \\
 & Add               & 96.82 & 83.85 & 91.22 & 90.11 & 92.36 \\ 
 & Subtract             & 96.83 & 83.85 & 91.21 & 89.70 & 92.78 \\
\hline
\multirow{6}{*}{ResNet50} 
 & LE    & 96.78 & 83.61 & 91.08 & 89.74 & 92.45 \\
 & CE  & 96.72 & 83.29 & 90.89 & 89.30 & 92.53 \\
 & SE  & 96.76 & 83.53 & 91.03 & 89.80 & 92.29 \\
\cline{2-7}
 & Concat            & 96.51 & 82.29 & 90.28 & 88.57 & 92.06 \\
 & Add               & 96.48 & 82.09 & 90.17 & 88.12 & 92.31 \\
 & Subtract             & 96.57 & 82.63 & 90.49 & 88.97 & 92.06 \\
\hline
\multirow{6}{*}{EfficientNet-B4} 
 & LE    & 96.84 & 83.93 & 91.26 & 90.15 & 92.40 \\
 & CE  & 96.74 & 83.33 & 90.91 & 89.08 & 92.81 \\
 & SE  & 96.76 & 83.53 & 91.03 & 89.80 & 92.29 \\
\cline{2-7}
 & Concat            & 96.60	& 82.72	& 90.54	& 88.89	& 92.26 \\
 & Add               & 96.63	& 82.97	& 90.69	& 89.61	& 91.81 \\
 & Subtract          & 96.72	& 83.22	& 90.84	& 88.82	& 92.96 \\

\hline
\end{tabular}
\end{table*}

\begin{table*}[htbp]
\centering
\caption{Test Results Of Different Backbones Combined With Various Feature Exchange Mechanisms On The CDD Dataset}
\label{tab:cdd_backbone}
\begin{tabular}{l c c c c c c}
\hline
\textbf{Backbone} & \textbf{Exchange Method} & \textbf{OA} & \textbf{IoU} & \textbf{F1} & \textbf{Rec} & \textbf{Prec} \\
\hline
\multirow{6}{*}{SwinT V2} 
 & LE    & 99.64 & 97.11 & 98.53 & 98.44 & 98.63 \\
 & CE  & 99.59 & 96.75 & 98.35 & 98.36 & 98.34 \\
 & SE  & 99.55 & 96.40 & 98.17 & 98.02 & 98.32 \\
\cline{2-7}
 & Concat            & 99.63 & 97.05 & 98.50 & 98.58 & 98.43 \\
 & Add               & 99.63 & 97.01 & 98.48 & 98.42 & 98.54 \\
 & Subtract             & 99.56	& 96.49	& 98.22	& 98.11	& 98.32 \\

\hline
\multirow{6}{*}{ResNet50} 
 & LE    & 99.27 & 94.22 & 97.02 & 96.51 & 97.55 \\
 & CE  & 99.22 & 93.84 & 96.82 & 96.40 & 97.24 \\
 & SE  & 99.05 & 92.50 & 96.11 & 95.34 & 96.88 \\
\cline{2-7}
 & Concat            & 99.26	& 94.16	& 96.99	& 96.56	& 97.42 \\
 & Add               & 99.14 & 93.23 & 96.50 & 95.92 & 97.09 \\
 & Subtract             & 99.23 & 93.93 & 96.87 & 96.76 & 96.99 \\
\hline
\multirow{6}{*}{EfficientNet-B4} 
 & LE    & 99.28 & 94.30 & 97.06 & 96.65 & 97.48 \\
 & CE  & 99.22 & 93.86 & 96.83 & 96.91 & 96.76 \\
 & SE  & 99.03 & 92.42 & 96.06 & 95.87 & 96.26 \\
\cline{2-7}
 & Concat            & 99.28	& 94.29	& 97.06	& 96.89	& 97.23 \\
 & Add               & 99.15	& 93.32	& 96.54	& 96.35	& 96.73 \\
 & Subtract             & 99.17	& 93.48	& 96.63	& 96.38	& 96.88 \\

\hline
\end{tabular}
\end{table*}

\begin{table*}[htbp]
  \centering
  \caption{Comparison of Params (M) and FLOPs (G) between SEED Architecture and Fusion Methods}
  \label{tab:params_gflops}
  \begin{tabular}{l|ccc}
    \hline
    \textbf{Method}
      & \multicolumn{3}{c}{\textbf{Params (M)/FLOPs (G)}} \\
    \cline{2-4}
      & \textbf{SwinTv2} & \textbf{EfficientNet-B4} & \textbf{ResNet50} \\
    \hline
    SEED (LE)               & 74.18/109.51 & 25.53/61.92  & 33.22/77.61 \\
    SEED (CE)               & 74.18/109.51 & 25.53/61.92  & 33.22/77.61 \\
    SEED (SE)               & 74.18/109.51 & 25.53/61.92  & 33.22/77.61 \\
    \hline
    SEED (Single Decoder)   & 74.18/84.82  & 25.53/34.78  & 33.22/49.60 \\
    \hline
    Add                     & 74.18/84.82  & 25.53/34.78  & 33.22/49.60 \\
    Subtract                & 74.18/84.82  & 25.53/34.78  & 33.22/49.60 \\
    \hline
    Concat                  & 74.68/85.33  & 25.71/34.91  & 34.21/50.61 \\
    \hline
  \end{tabular}
\end{table*}

\section{Discussion}
\subsection{Why is feature exchange effective for change detection?}

Prior work has explored feature exchange to enhance cross-temporal interaction, but typically kept an explicit differencing module and treated exchange as auxiliary. In contrast, SEED relies on exchange alone and removes explicit differencing. Empirically, a simple encoder-decoder with exchange already matches or surpasses stronger pipelines with handcrafted differences. This raises a key question: \emph{why can exchange alone suffice?}

\subsubsection{Information preservation under pixel consistency.}
In SEED, each exchange is a permutation (orthogonal) operator applied along the layer, channel, or spatial axis. Under pixel consistency (i.e., correspondence of paired pixels is preserved), such permutations are invertible and isometric, thus they preserve mutual information and Bayes-optimal risk with respect to the change label. Consequently, exchange injects cross-temporal conditioning without discarding task-relevant information, unlike non-invertible fusions (addition, subtraction, or concatenation followed by reduction).

\subsubsection{Intuitive exposure of disagreement cues.}
Channel and layer exchanges permute feature channels across times; spatial exchange permutes stripes (rows/columns) across times. Although these operate on deep features (not raw RGB), the effect is to expose cross-temporal inconsistencies in-place. As illustrated in Figure~\ref{fig:seed_exchange_vis}, applying channel or spatial exchange to examples with large changes makes the changed regions visually salient, which simplifies supervised learning of the change mask.

\subsubsection{A minimal toy experiment.}
We further validate this intuition on LEVIR-CD with a minimal segmentation model: an EfficientNet-B3 encoder and a residual, layer-wise decoder. Given bi-temporal inputs \(x_A\) and \(x_B\), we form exchanged inputs \(x'_{AB}\) and \(x'_{BA}\) via channel or spatial exchange. We train \emph{one} set of encoder--decoder weights on both \(x'_{AB}\) and \(x'_{BA}\) using the standard binary change mask as supervision. As shown in Figures~\ref{fig:channel_training} and \ref{fig:spatial_training}, both exchange types converge reliably and fit the change regions well, even without any explicit differencing branch.

\subsubsection{Takeaway.}
Exchange is a parameter-free, information-preserving way to couple the two times while keeping per-time semantics intact. It makes disagreement cues easier to learn, which explains why SEED---a Siamese encoder--exchange--decoder without fusion/differencing heads---achieves strong results. These observations are consistent with our formal analysis of exchange as an orthogonal permutation and its mutual-information invariance.

\begin{table*}[htbp]
    \centering
    \caption{Performance Metrics for Random Exchange Methods on Change Detection Datasets}
    \makebox[\textwidth][c]{
    \label{tab:random_exchange_results}
    \begin{tabular}{l c c c c c c}
        \hline
        \textbf{Dataset} & \textbf{Random Exchange} & \textbf{OA} & \textbf{IoU} & \textbf{F1} & \textbf{Rec} & \textbf{Prec} \\
        \hline
        \multirow{3}{*}{SYSU-CD} 
            & LE   & 91.09 & 67.49 & 80.59 & 78.40 & 82.90 \\
            & CE & 91.32 & 67.11 & 80.32 & 75.10 & 86.31 \\
            & SE  & 91.31 & 67.95 & 80.91 & 78.15 & 83.88 \\
        \hline
        \multirow{3}{*}{LEVIR-CD} 
            & LE   & 99.10 & 83.43 & 90.96 & 89.31 & 92.69 \\
            & CE & 99.24 & 85.97 & 92.45 & 91.04 & 93.92 \\
            & SE & 99.21 & 85.46 & 92.16 & 90.83 & 93.53 \\
        \hline
        \multirow{3}{*}{PX-CLCD} 
            & LE   & 99.25 & 94.30 & 97.07 & 96.70 & 97.43 \\
            & CE & 99.30 & 94.73 & 97.30 & 97.51 & 97.08 \\
            & SE & 99.32 & 94.83 & 97.35 & 97.12 & 97.57 \\
        \hline
        \multirow{3}{*}{WaterCD} 
            & LE   & 96.93 & 84.29 & 91.48 & 89.89 & 93.12 \\
            & CE & 96.86 & 84.04 & 91.33 & 90.22 & 92.46 \\
            & SE & 96.80 & 83.81 & 91.19 & 90.44 & 91.96 \\
        \hline
        \multirow{3}{*}{CDD} 
            & LE   & 99.44 & 95.55 & 97.73 & 97.61 & 97.85 \\
            & CE & 99.53 & 96.26 & 98.09 & 97.93 & 98.26 \\
            & SE & 99.50 & 96.02 & 97.97 & 97.74 & 98.20 \\
        \hline
    \end{tabular}
    }
\end{table*}

In the preceding experiments, we showed that a simple encoder--decoder equipped with feature exchange can already learn change regions effectively. To further assess the \emph{flexibility} of exchange, we trained with a stochastic exchange policy and evaluated with a fixed policy. Concretely, during training we applied random layer/channel/spatial exchange, at each iteration, different subsets of layers, channels, or spatial stripes were swapped across times. During validation and inference, the exchange pattern was held fixed to isolate the effect of training-time stochasticity.

Using SEED with a Swin Transformer~V2 backbone, the randomized setting yields competitive results (Table~\ref{tab:random_exchange_results}). For example, on LEVIR-CD, random spatial exchange attains an IoU of 85.97\%. On CDD, it reaches 96.26\%. These outcomes indicate that, under pixel consistency, SEED is insensitive to the precise exchange pattern and can leverage stochastic exchange without degrading accuracy. This robustness is consistent with our analysis of exchange as an information-preserving permutation.

\begin{table*}[htbp]
\centering
\caption{Quantitative Results of Different SEG2CD Algorithms on the LEVIR-CD Dataset}
\label{tab:seg2cd_levir_cd_results}
\makebox[\textwidth][c]{
\begin{tabular}{l l l c c c c c}
\hline
\textbf{Type} & \textbf{Model} & \textbf{Backbone} & \textbf{OA} & \textbf{IoU} & \textbf{F1} & \textbf{Rec} & \textbf{Prec} \\
\hline
\multirow{5}{*}{RS}
    & UNetFormer \cite{wang2022unetformer}     & ResNet18              & 99.12 & 83.62 & 91.08 & 87.92 & 94.47 \\
    & A2FPN \cite{li_a2-fpn_2022}         & ResNet18              & 99.09 & 83.33 & 90.91 & 89.51 & 92.35 \\
    & MANet \cite{li_multiattention_2022}         & ResNet50              & 99.19 & 84.87 & 91.81 & 88.89 & 94.94 \\
    & AFENet \cite{gao_adaptive_2025}        & ResNet18              & 99.22 & 85.65 & 92.27 & 90.88 & 93.70 \\
    & CMTFNet \cite{wu_cmtfnet_2023}       & ResNet50              & 99.12 & 83.59 & 91.06 & 88.05 & 94.28 \\
\hline
\multirow{3}{*}{CV}
    & DeepLabV3+ \cite{chen2018encoder}  & Xception65            & 99.18 & 84.76 & 91.75 & 89.19 & 94.47 \\
    & SegFormer \cite{xie_segformer_2021}      & MixVisionTransformer  & 99.10 & 83.35 & 90.92 & 88.83 & 93.12 \\
    & UPerNet \cite{xiao_unified_2018}       & ResNet50              & 99.22 & 85.39 & 92.12 & 89.69 & 94.69 \\
\hline
\end{tabular}
}
\end{table*}

\begin{table*}[htbp]
\centering
\caption{Quantitative Results of Different SEG2CD Algorithms on the SYSU-CD Dataset}
\label{tab:seg2cd_sysu_cd_results}
\makebox[\textwidth][c]{
\begin{tabular}{l l l c c c c c}
\hline
\textbf{Type} & \textbf{Model} & \textbf{Backbone} & \textbf{OA} & \textbf{IoU} & \textbf{F1} & \textbf{Rec} & \textbf{Prec} \\
\hline
\multirow{5}{*}{RS}
    & UNetFormer \cite{wang2022unetformer}     & ResNet18              & 91.73 & 68.82 & 81.53 & 77.43 & 86.09 \\
    & A2FPN \cite{li_a2-fpn_2022}          & ResNet18              & 91.19 & 66.55 & 79.92 & 74.36 & 86.37 \\
    & MANet \cite{li_multiattention_2022}          & ResNet50              & 90.95 & 65.98 & 79.51 & 74.45 & 85.29 \\
    & AFENet \cite{gao_adaptive_2025}         & ResNet18              & 92.03 & 69.98 & 82.34 & 78.73 & 86.29 \\
    & CMTFNet \cite{wu_cmtfnet_2023}        & ResNet50              & 90.76 & 67.90 & 80.88 & 82.84 & 79.01 \\
\hline
\multirow{3}{*}{CV}
    & DeepLabV3+ \cite{chen2018encoder}  & Xception65            & 91.67 & 68.53 & 81.33 & 76.93 & 86.26 \\
    & SegFormer \cite{xie_segformer_2021}      & MixVisionTransformer  & 91.36 & 67.87 & 80.86 & 77.42 & 84.62 \\
    & UPerNet \cite{xiao_unified_2018}        & ResNet50              & 91.50 & 68.31 & 81.17 & 77.68 & 85.00 \\
\hline
\end{tabular}
}
\end{table*}

\subsection{Applicability: converting segmentation to change detection (SEG2CD)}

SEG2CD is not a primary contribution of this paper; rather, it serves as evidence that exchange alone is sufficient to transfer encoder-decoder segmentation models to change detection with minimal engineering. Concretely, we duplicate the encoder and decoder into a Siamese topology with \emph{shared} weights, and insert a \emph{parameter-free} layer exchange between the Siamese encoder and decoder. No other architectural changes or hyper-parameter tuning are introduced.

This minimal recipe unifies semantic segmentation and change detection in the SEED view: explicit differencing is unnecessary; instead, change cues are learned directly from exchanged bi-temporal features by the shared Siamese encoder-decoder.

We evaluate this conversion on both remote-sensing (RS) and natural-scene (CV) segmentation models (Tables~\ref{tab:seg2cd_levir_cd_results} and \ref{tab:seg2cd_sysu_cd_results}). Despite their small backbones and the absence of task-specific tuning, SEG2CD variants achieve competitive OA/IoU/F1 on LEVIR-CD and SYSU-CD. For example, on LEVIR-CD, \textit{AFENet (ResNet-18)} reaches 92.27\% F1 and 85.65\% IoU, while \textit{DeepLabV3+ (Xception-65)} attains 91.75\% F1 and 84.76\% IoU. On SYSU-CD, \textit{AFENet (ResNet-18)} yields 82.34\% F1 and 69.98\% IoU, and \textit{DeepLabV3+ (Xception-65)} obtains 81.33\% F1 and 68.53\% IoU. These results indicate that, under SEED, standard segmentation architectures can be repurposed as strong CD baselines by inserting exchange alone. Therefore, these experimental results further prove the effectiveness of the SEED architecture and truly build a good bridge from the semantic segmentation model to the change detection model.

\begin{table}[!htb]
  \centering
  \caption{Quantitative results of single-decoder inference for the SEED architecture based on the IoU metric. Layer Exchange: LE; Channel Exchange: CE; Spatial Exchange: SE; A and B represent the two decoder branches; $\Delta$ indicates the mean difference between the single-decoder inference results (A and B) and the Siamese decoder inference results.}
  \label{tab:inference_single_decoder}
  \begin{tabular}{l c c c c c}
    \toprule
    \textbf{Dataset} & \textbf{Method} & \textbf{A} & \textbf{B} & \textbf{Siamese} & \textbf{$\Delta$} \\
    \midrule
    \multirow{3}{*}{SYSU-CD}
      & LE & 69.96 & 70.27 & 70.33 & $-0.22$ \\
      & CE & 70.65 & 70.79 & 70.91 & $-0.19$ \\
      & SE & 68.66 & 68.72 & 69.05 & $-0.36$ \\
    \midrule
    \multirow{3}{*}{LEVIR-CD}
      & LE & 86.09 & 86.01 & 86.25 & $-0.20$ \\
      & CE & 85.40 & 86.16 & 86.03 & $-0.25$ \\
      & SE & 85.57 & 86.38 & 86.24 & $-0.27$ \\
    \midrule
    \multirow{3}{*}{PX-CLCD}
      & LE & 95.24 & 95.14 & 95.34 & $-0.15$ \\
      & CE & 95.11 & 95.53 & 95.50 & $-0.18$ \\
      & SE & 94.50 & 94.71 & 94.86 & $-0.26$ \\
    \midrule
    \multirow{3}{*}{WaterCD}
      & LE & 84.54 & 84.60 & 84.64 & $-0.07$ \\
      & CE & 84.28 & 84.23 & 84.43 & $-0.18$ \\
      & SE & 83.88 & 83.83 & 84.06 & $-0.21$ \\
    \midrule
    \multirow{3}{*}{CDD}
      & LE & 96.91 & 96.92 & 97.11 & $-0.20$ \\
      & CE & 96.48 & 96.48 & 96.75 & $-0.27$ \\
      & SE & 96.07 & 96.05 & 96.40 & $-0.34$ \\
    \bottomrule
  \end{tabular}
\end{table}

\subsection{Exploring the Applications of SEED Paradigm in Change Detection Tasks}  
\subsubsection{SEED in Lightweight Change Detection}  
Before the introduction of the SEED paradigm, change detection models had not moved beyond the difference feature construction module. Simple difference feature construction methods (Fusion methods) like Concat, Add, and Subtract do not provide an advantage in terms of accuracy. More complex difference feature construction methods naturally come with a higher computational cost. The SEED paradigm replaces the traditional difference feature construction module with a parameter-free feature exchange mechanism, completely discarding the limitations of difference feature construction, and significantly improving the advantages of lightweight design. Specifically, the SEED paradigm adopts a Siamese encoder-exchange-decoder structure, where both the Siamese encoder and decoder parameters are shared, significantly reducing the parameter count and memory usage of the change detection model.

To this end, we computed the parameters based on the SEED paradigm and difference feature construction methods, as shown in Table~\ref{tab:params_gflops}. In terms of the params metric, the SEED paradigm is identical to Fusion methods. However, due to SEED having a dual-decoder structure, its FLOPs metric is higher. But if the inference process of the SEED paradigm only uses a single decoder, the FLOPs is identical to that of Fusion methods, as shown in SEED(Single Decoder) in Table~\ref{tab:params_gflops}.

Upon further reflection on the SEED paradigm, we identified a major advantage: the dual-path encoder-decoder structure is essentially equivalent. Therefore, both decoders in the SEED paradigm can independently produce inference results. To demonstrate this, we conducted the experiment shown in Table~\ref{tab:inference_single_decoder}, which shows that a change detection model based on the SEED paradigm can achieve good accuracy using only a single branch for inference. This experiment addresses the disadvantage of the SEED paradigm in terms of FLOPs compared to Fusion methods. Additionally, the SEED paradigm is very simple and highly compatible with semantic segmentation models or image classification models. As a result, the SEED paradigm can seamlessly integrate with various lightweight semantic segmentation models or lightweight backbone networks, enabling more flexible designs for lightweight change detection solutions.

\subsubsection{SEED in Self-Supervised Change Detection}  
In recent years, self-supervised learning algorithms have seen significant development in the field of computer vision. However, compared to the general field of computer vision, remote sensing images face domain-specific challenges due to the differences between natural scene images and remote sensing images. As a result, the remote sensing field requires more suitable pre-trained models. Compared to image segmentation and image classification, there is a lack of universal pre-trained models for change detection algorithms. The SEED paradigm introduces the first pure encoder-decoder change detection framework. As a result, compared to other change detection models that integrate difference computation modules, the SEED paradigm can be combined with Mask Autoencoder (MAE) ~\cite{he_masked_2021} to construct a dual-input, dual-output self-supervised model. Therefore, a self-supervised MAE model combined with the SEED paradigm can take dual-temporal image inputs, using feature exchange to swap tokens between the dual-temporal images. Finally, the Siamese decoders can be used to reconstruct the original images. Thus, by combining the SEED paradigm, a pre-trained model specifically for change detection algorithms can be constructed within the MAE framework.

\begin{table}[!htb]
\centering
\caption{Robustness to systematic pixel misregistration (IoU, \%). We shift the second temporal image by $N$ pixels (right shift) for train/val/test consistently. Results are obtained with layer exchange.}
\label{tab:shift_robustness}
\begin{tabular}{lccccc}
\toprule
Dataset & 0 & 2 & 4 & 6 & 8 \\
\midrule
LEVIR-CD & 86.25 & 85.30 & 84.23 & 83.35 & 83.23 \\
SYSU-CD  & 70.33 & 70.04 & 67.09 & 67.35 & 66.26 \\
PX-CLCD  & 95.34 & 95.14 & 95.34 & 95.03 & 95.04 \\
WaterCD  & 84.64 & 80.17 & 75.96 & 72.30 & 70.20 \\
CDD      & 97.11 & 95.49 & 96.13 & 96.41 & 96.37 \\
\bottomrule
\end{tabular}
\end{table}

\begin{figure}[htbp]
    \centering
    \includegraphics[width=0.8\textwidth]{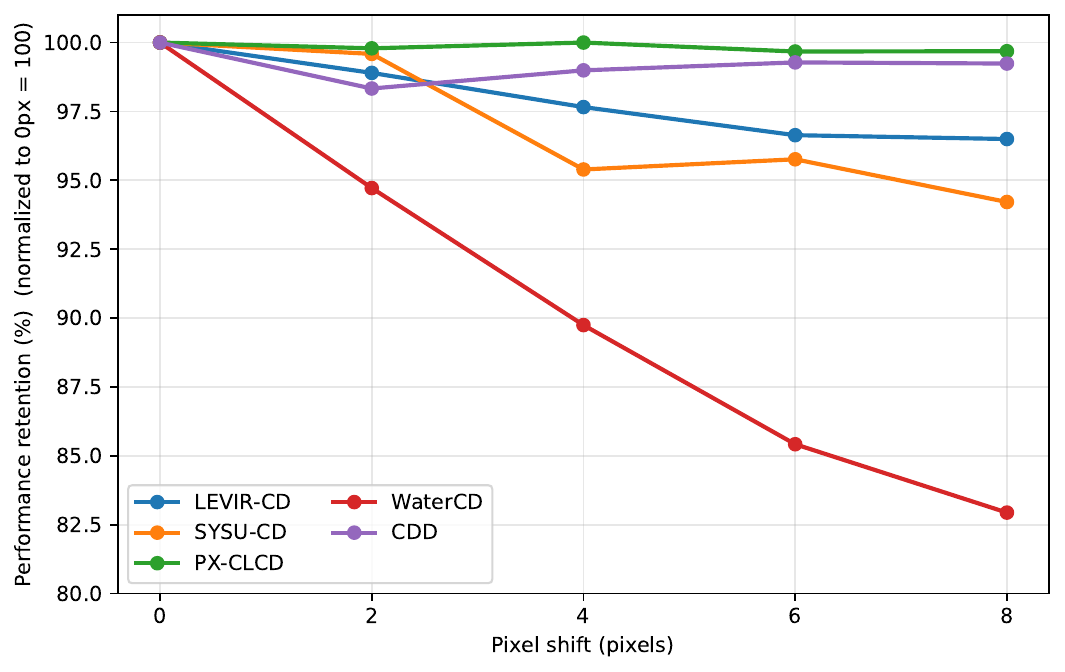} 
    \caption{Performance retention under systematic pixel misregistration. The second temporal image is shifted right by $N\in\{0,2,4,6,8\}$ pixels for all train/val/test splits. Retention is normalized to the $N{=}0$ setting (set to 100), and results are obtained with layer exchange.}
    \label{fig:shift_retention}
\end{figure}

\subsection{Robustness to Systematic Pixel Misregistration}

In practical remote sensing scenarios, perfectly co-registered bi-temporal images are not always available. Small global shifts may remain after preprocessing. Such shifts violate the strict pixel-consistency assumption and often make explicit differencing methods overly sensitive to local misalignment, thereby producing spurious ``change'' responses around object boundaries. Therefore, we conduct experiments to examine whether SEED remains effective when the two temporal inputs are not perfectly aligned, and whether the \emph{layer exchange} mechanism can tolerate such perturbations to some extent. This protocol evaluates trainability and performance retention under a consistent misregistration regime, rather than zero-shot robustness from aligned training to misaligned testing.

To simulate this situation under controlled conditions, we construct shifted variants by translating the second-temporal image to the right by a fixed number of pixels for all five benchmark datasets. Specifically, we consider shift magnitudes $N \in \{0, 2, 4, 6, 8\}$, and apply the same shift consistently to all bi-temporal pairs in the training, validation, and test sets. We then train SEED independently under each shift magnitude using the layer-exchange strategy, and evaluate the final performance on the correspondingly shifted test set.

The results in Table~\ref{tab:shift_robustness} show that SEED exhibits a gradual performance degradation as the misregistration magnitude increases. On LEVIR-CD, the IoU decreases from 86.25\% at $N{=}0$ to 83.23\% at $N{=}8$ (a drop of 3.02 points). On SYSU-CD, the IoU decreases from 70.33\% to 66.26\% (a drop of 4.07 points). For PX-CLCD and CDD, the performance remains highly stable across all shift magnitudes, with IoU variations of approximately 0.3, indicating that exchange-based coupling is not sensitive to small global misalignment on these datasets. WaterCD exhibits a larger drop (from 84.64\% to 70.20\%), which may be attributed to its 10\,m spatial resolution: an $N$-pixel shift corresponds to a real-world displacement of $N \times 10$\,m. Even so, the model still retains meaningful predictive capability at $N{=}8$.

To enable clearer comparison across datasets with different absolute IoU scales, Fig.~\ref{fig:shift_retention} further reports the normalized \emph{performance retention} by setting the $N{=}0$ performance to 100. The curves reveal a consistent trend: SEED maintains high retention under moderate misregistration, with near-flat behavior on PX-CLCD and CDD, and only mild degradation on LEVIR-CD and SYSU-CD. In contrast, WaterCD shows the steepest decline as $N$ increases, highlighting that datasets dominated by low spatial resolution are inherently more sensitive to pixel-level shifts. Overall, both the absolute IoU in Table~\ref{tab:shift_robustness} and the normalized retention in Fig.~\ref{fig:shift_retention} support the same conclusion: even when systematic misregistration partially violates pixel consistency, the layer-exchange mechanism keeps SEED trainable and maintains competitive accuracy, rather than failing catastrophically.

We attribute this behavior to the fact that layer exchange couples the two temporal streams at the feature level, without relying on handcrafted pixel-wise differencing. Even when pixel correspondence is partially invalid, exchange encourages the shared Siamese encoder--decoder to rely more on semantic and contextual cues aggregated in deep features, rather than treating local misalignment as direct evidence of change. As a result, SEED remains effective under 2--8 pixel shifts across multiple datasets, providing practical robustness in scenarios where perfect co-registration cannot be guaranteed.

\section{Conclusion}
We introduced SEED, a Siamese encoder--exchange--decoder paradigm for change detection. SEED removes explicit differencing and uses a parameter-free exchange plus shared encoders/decoders to learn change cues directly from bi-temporal features. We formalized exchange as a permutation operator and, under pixel consistency, showed it preserves mutual information and Bayes-optimal risk, explaining why common fusion/differencing (addition, subtraction, concatenation with reduction) can be information-losing. Across five datasets and three backbones, SEED matches or surpasses strong baselines with a simple and interpretable design; ablations confirm the robustness of layer, channel, spatial, and random exchanges. From a lightweight perspective, single-decoder inference recovers fusion-level FLOPs with comparable accuracy. We hope SEED offers a clear, flexible recipe for future change detection systems, including self-supervised pretraining tailored to bi-temporal data.

\bibliographystyle{unsrtnat}
\bibliography{references} 

\end{document}